\pdfoutput=1

\documentclass{article}

\usepackage{authblk}
\usepackage{stmaryrd}
\usepackage{pslatex}
\usepackage{apacite}
\usepackage{linguex}
\usepackage{xcolor}
\usepackage{tabularx}
\usepackage{multirow}
\usepackage{pbox}
\usepackage{colortbl}
\usepackage{fullpage}
\usepackage{natbib}
\usepackage[fleqn]{amsmath}

\newcommand{\sem}[1]{\ensuremath{\llbracket #1 \rrbracket}}
\newcommand{\pocite}[1]{\citeauthor{#1}'s \citeyear{#1}}

\usepackage{graphicx}

\usepackage{caption}
\usepackage{subcaption}

\usepackage{enumitem}

\renewcommand{\phi}{\varphi} 
\newcommand{\utt}[1]{\textit{#1}}

\title{Exhaustivity and anti-exhaustivity in the RSA framework:\\ Testing the effect of prior beliefs%
\thanks{We would like to thank Leon Bergen, Lucas Champollion, Judith Degen, Danny Fox, Michael Franke, Roni Katzir, MH Tessler for precious discussion, as well as the audiences of the LPCS research seminar at the University of Latvia, the System analysis group at Vilnius University, the NYU semantics group, the ESSLLI workshop ``Approaches to implicature: Rational choice and/or exhaustification'', the LINGUAE seminar at ENS Paris, and the semantics research seminar at UCL. We gratefully acknowledge funding from the
Agence Nationale de la Recherche (FrontCog Project ANR-17-EURE-0017 and ProbaSem project ANR-19-CE28-0004-01), the Mind Brain Behavior Interfaculty Initiative at Harvard, and the Research Council of Lithuania and European Social Fund under Measure 09.3.3-LMT-K-712.}
}
 
\author[1]{\textbf{Alexandre Cremers}}
\author[2]{\textbf{Ethan G.\ Wilcox}}
\author[3]{\textbf{Benjamin Spector}}

\affil[1]{Vilnius University, Filologijos Fakultetas}
\affil[2]{Harvard University}
\affil[3]{Institut Jean Nicod, D\'epartement d'\'etudes cognitives, ENS, EHESS, PSL University, CNRS}

\begin{document}

\maketitle

\begin{abstract}
During communication, the interpretation of utterances is sensitive to a listener's probabilistic prior beliefs, something which is captured by one currently influential model of pragmatics, the Rational Speech Act (RSA) framework. In this paper we focus on cases when this sensitivity to priors leads to counterintuitive predictions of the framework. Our domain of interest is exhaustivity effects, whereby a sentence such as \textit{Mary came} is understood to mean that \emph{only} Mary came. We show that in the baseline RSA model, under certain conditions, \emph{anti-exhaustive} readings are predicted (e.g., \textit{Mary came} would be used to convey that both Mary and Peter came). The specific question we ask is the following: should exhaustive interpretations be derived as purely \emph{pragmatic} inferences (as in the classical Gricean view, endorsed in the baseline RSA model), or should they rather be generated by an encapsulated \emph{semantic} mechanism (as argued in some of the recent formal literature)? To answer this question, we provide a detailed theoretical analysis of different RSA models---some of which include such an encapsulated semantic mechanism---, and evaluate these models against data obtained in a new study which tested the effects of prior beliefs on both production and comprehension, improving on previous empirical work. We found no anti-exhaustivity effects, but observed that message choice is sensitive to priors, as predicted by the RSA framework overall. The best models turn out to be those which include an encapsulated exhaustivity mechanism (in line with other studies which reached the same conclusion on the basis of very different data). We conclude that, on the one hand, in the division of labor between semantics and pragmatics, semantics plays a larger role than is often thought, but, on the other hand, the tradeoff between informativity and cost which characterizes all RSA models does play a central role for genuine pragmatic effects.
\end{abstract}

\section{Introduction}

The effect of linguistic utterances in context depends in part on the prior beliefs of speakers and hearers. Suppose that Mary and Peter are known to be a couple, who usually go out together, so that, upon learning that Mary came to a certain party to which Peter was also invited, one would assign a high probability to the possibility that Peter came as well. If someone is asked \textit{Did Mary come to the party?} and answers with \textit{she did}, the questioner would be warranted to infer that Peter probably came too. Furthermore, if the question is instead \textit{Did Peter come to the party?}, an answer such as \textit{I am not sure, but Mary did} would again suggest that it is quite likely that Peter came to the party.  Typically, though, pragmatic inferences go well beyond what can be predicted with such a simple model of linguistic interpretation. They also involve, for instance, reasoning about \textit{other sentences} that the speaker could have uttered given some assumptions about their communicative goals (\citealt{grice1975logic}). For instance, in response to the question \textit{Among Gloria, Adam, Fred and Sue, who attended the show today?}, an answer such as \textit{Adam did} tends to trigger the inference that the others did not, even if there is no expectation that what any of them does depends on what the others do. Such \textit{exhaustivity effects} can be accounted for in terms of Grice's maxim of quantity: if in fact both Adam and Gloria had attended the show, a knowledgeable speaker would say that rather than just talking about Adam. Alternatively, several arguments have been put forward in the literature according to which such exhaustivity effects are linguistically encoded and result from the presence, in the logical form of the sentence, of an \textit{exhaustivity operator} (see, e.g., \citealt{chierchia2012scalar}), whose meaning is, in first approximation, akin to that of \emph{only} (\emph{Only Mary did } entails that nobody else attended the show). In both types of accounts (Gricean and operator-based), a crucial ingredient is the idea that the sentence being interpreted (\textit{Mary did}) is enriched with the negation of some \textit{alternative} sentences (e.g, \textit{Peter and Mary did}).

Now,  in some situations, these two types of effects (effect of prior beliefs, exhaustivity effects) are pitted against each other (for a recent discussion of this, see \citealp{wilcox2019,schreiber2021narrow}). Consider again the case where Peter and Mary are known to typically go out together, and someone asks \textit{Among Peter, Mary, Fred and Sue, who attended the show today?}.  What should we expect an answer such as \emph{Mary did} to convey? On the one hand, given prior probabilistic knowledge, one should be led to conclude that Peter probably attended the show. On the other hand, the answer might trigger an exhaustive interpretation, whereby one would conclude the opposite, namely that only Mary attended the show.

{Our goal in this paper is to investigate the relationship between prior beliefs and exhaustivity effects, working within a pragmatic modeling framework, the Rational Speech Act framework \citep{frank2012predicting, goodman2013knowledge}. By modeling pragmatics as joint probabilistic reasoning between speaker/listeners layers, the RSA makes it possible to incorporate prior probabilistic knowledge into both message interpretation and message choice. We start at the theoretical level, presenting a detailed comparison of nine different variants of the RSA framework and an analytic description of the conditions under which each will produce exhaustivity effects. We demonstrate that, in addition to exhaustivity, many of the model variants tested are predicted to produce \textit{anti-exhaustivity} effects; a theoretical prediction distinctive to the RSA framework in which under certain conditions a sentence such as \emph{Mary came} ends up conveying, for instance, that both Peter and Mary came.\footnote{\citet{degen2015wonky}, which we will discuss in detail in section \ref{sec:wRSA}, as well as \citet{wilcox2019}, \citet{schreiber2021narrow}, and \citet{fox2021notes}, have pointed out before us that the predictions of the RSA model (and related frameworks in the case of of \citealt{fox2021notes}) are extremely sensitive to priors, in a way that is sometimes problematic.} To evaluate the model variants, we deploy an experimental paradigm that assesses the effect of prior beliefs on the interpretation and production of sentences, allowing us to test our RSA models as joint models of comprehension and production. We find strong evidence for exhaustivity effects, as well as that these effects can be influenced by prior knowledge; however, we find no evidence for anti-exhaustivity effects in human participants. Our joint data reveal that, for this pragmatic phenomenon, neither production nor comprehension can be both rational and well-informed, meaning that the best-fitting models are either rational but do not share some information between speakers/listeners, or models which depart from the rationality assumption for one of the agents. In addition, the lack of anti-exhaustivity effects in the data mean that the best-fitting RSA models are those which break the exhaustivity/anti-exhaustivity symmetry in the semantics via an encapsulated, non-pragmatic mechanism such as grammatical exhaustification.}

In order to {set up a crucial dimension along which different RSA variants can differ from each other}, we start in Section~\ref{sec:symmetry} with a discussion the so-called \emph{symmetry problem} for exhaustivity effects. We show that In contrast to `logic-based' approaches to exhaustivity, the baseline RSA model can predict not only exhaustive readings, but also, and quite surprisingly, anti-exhaustive readings, whereby sentences are enriched with (the positive form of) some of their alternatives.

\section{The symmetry problem: exhaustivity and anti-exhaustivity}
\label{sec:symmetry}

Consider an utterance of a certain sentence \utt{A}, in a context where the truth-values of two sentences \utt{A} and \utt{B} are relevant. Suppose the speaker is fully informed about \utt{A} and \utt{B}. If she believes that \utt{A} is true and \utt{B} is false, then the message  \utt{A and not B} is the most informative sentence she can use truthfully. If she believes that both \utt{A}  and \utt{B} are true, then \utt{A and B} is the most informative sentence she can use truthfully. So, if the speaker's choice of a message were only determined by informativity considerations, \utt{A} would never be used by a fully informed speaker, and should therefore give rise to the inference that the speaker does not know \utt{B}'s truth-value. Since, in fact, an utterance of \utt{A} typically gives rise to an exhaustivity effect, one must include a mechanism to break the symmetry between the two more informative messages \utt{A and B} and \utt{A and not B}. 

In the neo-Gricean tradition, the solution is based on the notion of \emph{alternatives}. It is assumed that \utt{A and B} is an alternative of \utt{A}, but that \utt{A and not B} is not. The exhaustive reading of a sentence is obtained by negating the alternatives that entail it (or, depending on the details, the alternatives that are not entailed by it). So in the case of \utt{A}, we add the negation of \utt{A and B}, and we get $A \wedge \neg B$. This approach is shared both by pragmatic accounts in the spirit of Grice (e.g., \citealp{sauerland2004scalar}) and by the so-called  `grammatical approach', based on exhaustivity operators. 

In recent years, several probabilistic, decision-theoretic models of pragmatics have proposed, among other things, a new formalization of exhaustivity effects (see, a.o, \citealt{parikh2001use,benz2005game,jager201293,franke2011quantity,rothschild2013game,frank2012predicting,goodman2013knowledge,bergen2016pragmatic}). We specifically focus here on the Rational Speech Act model (\citealt{frank2012predicting,goodman2013knowledge,bergen2016pragmatic}), which has become extremely influential.
While we will present the details of the framework in Section \ref{sec:base-RSA}, what is sufficient for the current discussion is that in this model, pragmatic reasoning is treated as probabilistic recursive reasoning between different `speaker' and `listener' layers. The speaker's choice of a message is governed by a tradeoff between informativity and cost: that is, the speaker tries to pick a message that will be maximally informative, in an information-theoretic sense, while minimizing the cost of the message. While in some specific applications, the set of messages that are considered in a given model is defined in terms of the classical neo-Gricean alternatives, neo-Gricean alternatives can in fact be entirely dispensed with, because cost can essentially serve the same purpose.\footnote{Of course, for any model to be tractable, it will include a finite number of messages, so in a sense every model makes certain commitments regarding alternative messages. In specific applications, typically, only a narrow range of messages is considered. This is sometimes done on the basis of the classical neo-Gricean notion of alternative, in which case there is no need to include a cost for messages (see \citealp[\S2.3.1]{franke2009signal} for an early discussion of whether to use cost or neo-Gricean alternatives to break symmetry in a probabilistic model of implicatures). But  the underlying assumption of an RSA model that includes a notion of \textit{cost} is that the availability of messages to the speaker depends on their cost and, in particular, that cost is responsible for breaking the symmetry between alternative messages. A model with cost of course includes a finite set of messages, but without relying on, say, Horn scales to solve the symmetry problem. The cost of a message might be viewed as determining the prior probability that a certain message will be used, independently of the speaker's communicative intent (the higher the cost, the less likely it is to be used), so that not including a message amounts to assigning it an infinite cost. In the conclusion, we come back to the implications of relying on the neo-Gricean notion of alternatives rather than on cost within the RSA framework. \label{alternativesfn}} Things work as follows. \utt{A and not B} is more costly than \utt{A and B}. For this reason, a speaker will typically be more likely to choose the less informative message \utt{A} when \utt{A} is true and \utt{B} is false than when both \utt{A} and \utt{B} are true. In both cases, the more informative message, which is either \utt{A and not B}  or \utt{A and B}, is more costly than \utt{A} itself, but the cost disadvantage is more pronounced in the case of \utt{A and not B}. As a result, the speaker has a stronger incentive to use the cheap but underinformative message \utt{A} when \utt{A} is true and \utt{B} is false than when both \utt{A} and \utt{B} are true.  Upon hearing \utt{A}, then, a Bayesian listener will infer that the former situation is more likely than the latter. 

However, while symmetry \emph{can} be broken by cost, it can also be broken by the other term of the tradeoff that governs message-choice, namely informativity, which is defined in probabilistic terms. Consider a context where the prior probability of \utt{B} {being true} given that \utt{A} {is true} is very high. In such a context, if the speaker believes that A is true and B is false, then the message \utt{A}, in terms of its literal meaning, is very bad at communicating the speaker's belief: upon learning \utt{A}, a literal listener will assign a very high probability to \utt{B} {being true}. Even if it does well on the cost side of the tradeoff, it does not do well at all on the informativity side. On the other hand, if the speaker believes that both \utt{A} and \utt{B} are true, then \utt{A} is a very good message to communicate her belief: by simply saying \utt{A}, the speaker is sure that the listener will assign a very high probability to \utt{B} {being true}, so the message \utt{A} is almost as good as the message \utt{A and B} in terms of informativity, and it is otherwise better in terms of cost.\footnote{In comparison, even when the conditional probability that, say, all $X$s $Y$ed given that some $X$s $Y$ed is very high, a \textit{some}-message can never be better than an \textit{all}-message to convey that all $X$s $Y$ed, because \textit{all} is no more costly than \textit{some}, but is more informative. While \cite{degen2015wonky} showed that the baseline RSA model predicts too strong an influence of priors in such case, a model where the speaker is fully rational would not have this problem (\textit{some} would never be used by a fully rational speaker to communicate \textit{all}). In contrast with this, even a fully rational speaker may prefer to use \utt{A} to convey that \utt{A and B} is true instead of \utt{A and B}, because \utt{A and B} is more costly than \utt{A}.}
 	
In other words, the speaker might now be \emph{more likely} to say \utt{A and not B} when  \utt{A and not B} is true than she is to say \utt{A and B} when \utt{A and B} is true. This means that she would be comparatively more likely to use \utt{A} when \utt{A and B} is true than when  \utt{A and not B} is. The symmetry between the two alternative messages \utt{A and B} and \utt{A and not B} is now broken by informativity considerations, in favor of \utt{A and not B}. For this reason, the baseline RSA model predicts cases of \emph{anti-exhaustivity}: when the priors are sufficiently biased, i.e.\ when {$P(B$ true$|A$ true$)$} is high enough, the pragmatic listener, after processing the message \utt{A}, will assign a higher probability to \utt{B} {being true} than what the priors by themselves warrant on the basis of the literal meaning of \utt{A}: \ $P(B\text{ true}| \text{ speaker said`}A\text{'})$ (the posterior probability assigned to \utt{B} being true after processing the message \utt{A}) will be higher than $P(B \text{ true} | A \text{ true})$ (the conditional prior belief).

Now, as has been previously noted \citep{degen2015wonky,javangula2019,wilcox2019,fox2021notes,schreiber2021narrow}, this prediction seems to be undesirable. Consider a context where Mary is known to be nearly always accompanied by Peter anywhere she goes. In the context of a question such as \utt{Did Mary come to the party?}, the answer \utt{Yes she did} {could plausibly} license the inference that Peter probably was there too.  In such a case, though, it is not part of the speaker's communicative intention to convey that Peter was there, given the underlying question, which is only about Mary. In contrast, if the question is \utt{Who came among Peter, Mary and Sue?}, the answer \utt{Mary did} {is very unlikely to be used, it seems to us,} by a speaker who wants to convey that Mary and Peter came but Sue didn't, and would not be so interpreted (see section~\ref{sec:empirical-background} for previous experimental results confirming this point).

{But while the baseline model is subject to potentially unwanted anti-exhaustivity effects, it is not the} only one model within the RSA framework, and different RSA models will make different predictions. For instance, \cite{degen2015wonky} address a closely related problem and offer a solution based on a refined RSA model, where the listener is, in a sense, uncertain about the priors that the speaker is assuming (we will discuss their proposal in details in section~\ref{sec:wRSA}). Let us also note that the baseline RSA model, despite making \emph{prima facie} undesirable anti-exhaustivity inferences, makes at the same time a plausible prediction about the influence of priors on message choice. Namely, it predicts that if $A \wedge \neg B$ is antecedently much less likely to be true than $A \wedge B$, a speaker who wants to convey $A \wedge \neg B$ will have an extra incentive to use the very explicit message \utt{A and not B} rather than simply say \utt{A}.  For instance, in the above situation, the speaker might be more likely to say \utt{Mary, but not Peter} than in a situation where there is no expectation that Peter and Mary would typically come to the party together (see \citealp[][\S3]{chemla2015probability} for a similar observation regarding \utt{some}/\utt{some but not all}).

The rest of this paper proceeds as follows: In section~\ref{sec:base-RSA} we present the baseline model, and provide an analytic description of when it produces both exhaustivity and anti-exhaustivity effects. In section~\ref{sec:other-models} we present alternative variants of the baseline RSA model, such as the `wonky worlds' model (\citealp{degen2015wonky}) as well as models in which sentences are potentially ambiguous, which makes it possible to treat exhaustive interpretations as being derived in the semantics: the Lexical Uncertainty model (\citealp{bergen2016pragmatic}),  \cite{degen2015wonky}, the Lexical Intentions model (\citealp{franke2020theory}), as well as another variant of the RSA framework that we call the `supervaluationist model' (based on \citealp{spector2017homogeneity}). In each case, we discuss whether these models can in principle predict anti-exhaustive readings and whether they predict an influence of priors on message choice. Finally, in Section~\ref{sec:experiment}, we will report the results of {an experiment probing both interpretation and production} and compare how well different models are able to account for the results.

\section{The baseline RSA model and anti-exhaustivity effects}
\label{sec:base-RSA}

\subsection{The base-line RSA model}

The RSA framework is  a game-theoretic and decision-theoretic model of pragmatics. Speakers are modeled as approximately rational speakers whose choice of a message is governed by a tradeoff between the \emph{informativity} (measured in information-theoretic terms) and the \emph{cost} of messages. Listeners are viewed as  Bayesian agents who update their beliefs about the world and the communicative intentions of the speaker based on the message that the speaker used. The model defines a sequence of listeners and speakers of increasing sophistication. Starting with a literal listener who simply conditionalizes her belief state (viewed as a probability distribution) on the meaning of the message, the model defines a first-level pragmatic speaker who believes she is talking to the literal listener, and chooses her message accordingly. It then defines a first-level pragmatic listener who uses Bayes' rule to update her beliefs, on the assumption that the received message was produced by the pragmatic speaker. An even more sophisticated (second-level) pragmatic speaker can then be defined as one who believes that she is speaking to the first-level pragmatic listener, and so on. 
 
The baseline RSA model includes: a set of possible world states (worlds for short); a probability distribution $P$ over world states which can be thought of as representing the listener's information state, known to the speaker, before she heard any message; a set of messages, each of which have a fixed meaning (a function from worlds to truth values). Each message $m$ has a \emph{cost} $c(m)$. Finally, the model includes a `rationality parameter' $\lambda$ (a positive real number) that regulates how rational the speaker is when choosing a message (the higher $\lambda$ is, the more likely it is that the speaker will choose the best message).

The \textit{literal listener}, $L_0$ interprets messages by conditionalizing {their distribution over world states}, $P$, with the proposition expressed by the literal meaning of the message. If $u$ is a message, we identify its meaning with the set of worlds in which it is true, and notate this set \sem{u}, and we notate $\sem{u}(w)$ the truth-value of $u$ in $w$ . $L_0$ is defined by the probability it assigns to a world $w$ after having processed a message $u$:


\ex. $ L_0(w | u) = P (w | \llbracket u \rrbracket) = \left\{\begin{array}{cl}
 0 & \text{if $w \not\in \llbracket u \rrbracket$,} \\
 \frac{P(w)}{P(\llbracket u \rrbracket)} & \text{if $w \in  \llbracket u \rrbracket$}
\end{array}\right.$\\ Equivalently (using the proportionality notation): $L_0(w | u) \propto P(w)\sem{u}(w)$

The 1st-level pragmatic speaker is characterized by a \textit{utility function} which assigns a utility to a  {message-world pair}, based on how the literal listener interprets $u$ and the cost of the message, notated $c(u)$:

\ex. $U_1(u,w) = \log(L_0(w|u)) - c(u)$

Importantly, this function expresses a tradeoff between informativity and cost. The first term is an increasing function of the probability assigned by $L_0$ to $w$ after having processed $u$, and so represents how effective $u$ at communicating $w$ to $L_0$.

The level-1 speaker is characterized by a function $S_1$ which determines how likely the speaker is to choose a message $u$ when she wants to communicate $w$, and is defined by the following equation:

\ex. $S_1(u|w)= \dfrac{\exp(\lambda U_1(u,w))}{\sum\limits_{u'} \exp\left(\lambda U_1(u',w)\right)}$

When $\lambda$ approaches infinity, this speaker approaches a fully rational speaker who, in world $w$ will choose a message which maximizes $U_1(u|w)$ with probability 1. When $\lambda$ is finite, the speaker is only approximately rational, and messages that yield greater utility have a grater probability of being used than less useful messages, but they are not certain to be used.

Then we can define a \textit{pragmatic listener} $L_1$ who believes that she is receiving a message from $S_1$ and uses Bayes' rule to update a probability distribution over worlds when processing a message $u$:

\ex. $L_1(w|u) = \dfrac{P(w)S_1(u|w)}{\sum\limits_{w'} P(w')S_1(u|w')}$

On this basis, higher-level speakers and listeners are defined recursively as follows:

\ex.
\a. $U_{n+1}(u|w) = \log(L_n(w|u)) - c(u)$ \label{utility}
\b. $S_{n+1}(u|w) = \dfrac{\exp(\lambda U_{n+1}(u,w))}{\sum\limits_{u'} \exp(\lambda U_{n+1}(u',w))}$
\c. $L_n(w|u) =  \dfrac{P(w)S_n(u|w)}{\sum\limits_{w'} P(w')S_n(u|w')}$

\subsection{The symmetry problem and the tradeoff between informativity and cost: exhaustivity and  anti-exhaustivity} 
\label{sec:anti-exh-RSA}

Coming back to exhaustivity, let us consider a very simple RSA model, with just two worlds, $w_a$ (in which \utt{A} is true but \utt{B} is false) and $w_{ab}$ (in which both \utt{A} and \utt{B} are true), and three messages: \utt{A}, $A \wedge B$ (true if both \utt{A} and \utt{B} are true) and $A \wedge \neg B$ (true if A is true and B is false). We set the cost of \utt{A} to 0, and we notate the costs of the two other messages $c_{A\wedge B}$ and $c_{A\wedge\neg B}$.  In such a model, \utt{A} is uninformative, as it is true in both worlds. Suppose the speaker is fully informed about the world. In order to communicate $w_a$, the speaker will choose between \utt{A} and $A \wedge \neg B$ (the message $A \wedge B$ is false in $w_A$, and so its utility will be infinitely negative). In order to communicate $w_{ab}$, the speaker will choose between \utt{A} and $A \wedge B$. The statements in \Next describe the precise conditions under which anti-exhaustivity effects are expected (see Appendix \ref{sec:RSA-antiexh-proof} for proofs):

\ex. Anti-exhaustivity \label{anti-exh}%
\vspace{-.5\baselineskip}\[\begin{array}{llcl}
\text{a.} & S_1(A|w_{ab}) > S_1(A \wedge B|w_{ab}) & \text{iff} & -\log(P(w_{ab}))~< ~c_{A\wedge B}\\
\text{{b}.} & L_1(w_{ab}|A) > P(w_{ab}) & \text{iff} & \log(P(w_{ab})) - \log(P(w_a)) > c_{A \wedge \neg B} - c_{A \wedge B}
\end{array}\]

\Last[a] expresses the condition for observing an anti-exhaustivity effect on the speaker side. It means that  in $w_{ab}$, the level-1 speaker will prefer to use the underinformative message \utt{A} rather than $A \wedge B$ if the information gain that would be brought by $A \wedge B$ (measured by $-\log(P(w_{ab}))$) is not worth its cost: this will happen when $w_{ab}$ has a very high prior probability, so that the speaker has no incentive to use a costly message to convey $w_{ab}$, given that the listener already assigns a high probability to  $w_{ab}$. \Last[b] expresses the condition for observing anti-exhaustivity on the listener side. This condition is  
the mathematically precise version of the fact that symmetry between alternative messages can be broken by informativity:  when the bias in favor of $w_{ab}$ compared to $w_a$ (measured by the {log-odds} $\log(P(w_{ab})) - \log(P(w_a))$) exceeds the cost disadvantage of $A \wedge \neg B$ compared to $A \wedge B$, \utt{A} is more likely to be used in $w_{ab}$ than in $w_a$, and as a result the listener tends to interpret \utt{A} as conveying $w_{ab}$.

\begin{figure}[h]
\centering
\includegraphics[width=.6\textwidth]{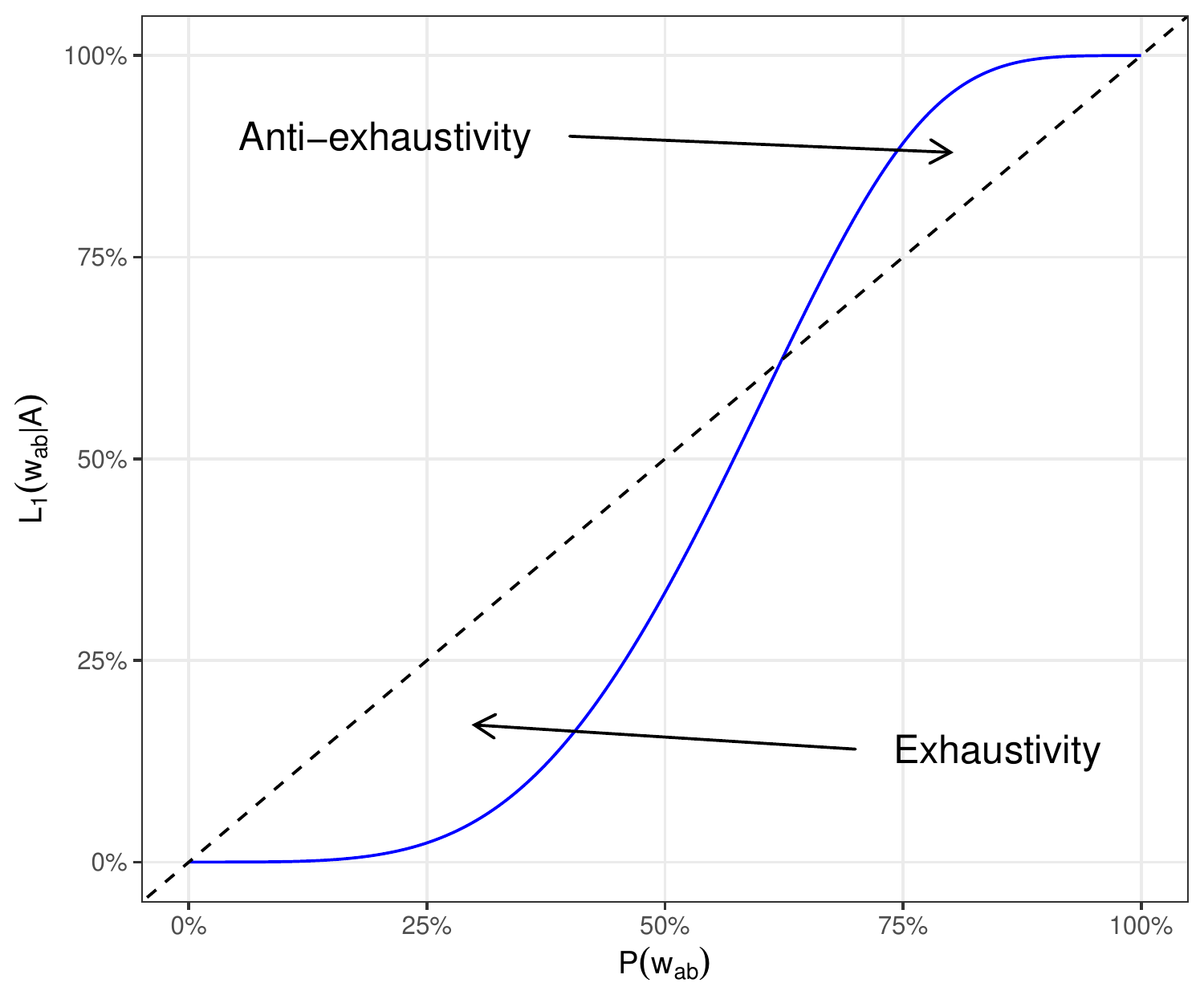}
\caption{Illustration of the anti-exhaustivity effect with high priors in baseline RSA ($\lambda=3$, $c_{A \wedge B}=.5$, $c_{A \wedge \neg B}=1$).}
\label{fig:antiexh-graph}
\end{figure}

To illustrate, Figure~\ref{fig:antiexh-graph} represents the predicted interpretation of \utt{A}  as a function of the prior probability of $w_{ab}$, for a certain choice of values for the parameters of the model (costs, rationality parameter). While anti-exhaustivity is a problematic prediction of the baseline RSA model, \Last[a] has a pendant \Next which appears to be a good prediction:

\ex. $S_1(A \wedge \neg B|w_a) > S_1(A |w_a)$ iff $- \log(P(w_{a})) >  c_{A \wedge \neg B}$ \label{prod}

In words: in $w_a$, the speaker will prefer the explicit and fully informative message $A \wedge \neg B$ over the less informative message \utt{A} if the prior probability of $w_{a}$ is low enough, so that the informativity gain brought by the fully informative message (measured by $-\log(P(w_{a}))$) exceeds its cost. \\
\\
 \noindent To sum up, the baseline RSA makes a number of interesting predictions when priors are such that $P(w_{ab})$ is much higher than $P(w_a)$:
 
 \begin{itemize}
 	\item  A speaker who believes $w_a$ is much more likely to say $A \wedge \neg B$ than in the case of non-biased priors.
 	\item A speaker who believes $w_{ab}$ will tend use the message \utt{A} rather than $A \wedge B$, even when the truth-value of both \utt{A} and \utt{B} are relevant (anti-exhaustivity on the speaker side).
 	\item The message \textit{A} will tend to be interpreted as conveying $w_{ab}$ (anti-exhaustivity on the listener side).
\end{itemize}

As has been established by previous studies \citep{javangula2019,wilcox2019,schreiber2021narrow}, and as we will find in our own experiments discussed in Section \ref{sec:experiment}, the last two predictions are problematic insofar as human subjects do not produce anti-exhaustivity effects. {The first one, however, is at least qualitatively correct.} In the next section, we turn to more sophisticated RSA models and ask whether they can solve the symmetry problem without giving rise to anti-exhaustivity{, but without losing the predicted effect of priors on the choice between \utt{A} and \utt{A$\wedge\neg$B} in the $w_a$ situation}.\footnote{A more complex model that also includes the messages \utt{B} and \utt{not B} makes the same qualitative predictions. In such a model, we need to include worlds in which $A$ is false but $B$ is true, since otherwise \utt{B} is simply equivalent to \utt{A and B}. Under uniform priors, in such a model, the speaker is even more likely to use \utt{A} in $w_a$ than in $w_{ab}$, because in $w_{ab}$ the message \utt{B}, which is no more costly that \utt{A}, could be used, while it could not be used in $w_a$ (while \utt{not B}, which is true in $w_a$, is more costly than $A$). Under biased priors (i.e.\ when $P(w_{ab}|\{w_a, w_{ab}\})$ is very high), \utt{A} can still be more likely to be used in $w_{ab}$ than in $w_a$, because \utt{A} would be a very bad message to use in $w_a$, while it is a quite good message to communicate $w_{ab}$. Furthermore, \utt{A}~ might be the best message to convey $w_{ab}$, because \utt{A and B} is more costly and only slightly more informative, while \utt{B}, while no more costly, is not necessarily more informative (since it is compatible with a world where only $B$ is true, whose prior probability could be high).}

\section{More sophisticated RSA models}
\label{sec:other-models}

As {mentioned earlier}, the baseline RSA model can be enriched in various ways. For instance, it can incorporate, as variables that speakers and listeners reason about, uncertainty about priors, uncertainty about what information is contextually relevant, or the Question Under Discussion (as in \citealt{kao2014nonliteral}), message ambiguity (as in the Lexical Uncertain model  of \citealp{bergen2016pragmatic}, the Lexical Intention model of \citealp{franke2020theory}, or \citeauthor{spector2017homogeneity}'s 2017 `supervaluationist' model described in section \ref{sec:spector}).\footnote{In all these cases, a fixed parameter of the baseline RSA becomes a probabilistic variable which is subject to Bayesian inference -- often then called a `lifted variable' -- or enters into the speaker's utility function (as in \citealt{spector2017homogeneity}).} In particular, the RSA framework is by itself compatible with the `grammatical approach' to exhaustivity, whereby sentences are systematically ambiguous   between a `plain' and an `exhaustified' meaning.  In this section, we describe and discuss the various models we will compare, namely \cite{degen2015wonky}, \cite{spector2017homogeneity}, the Lexical Uncertainty model \citep{bergen2016pragmatic} and the Lexical Intention model \citep{franke2020theory}, focusing in particular on whether these models predict antiexhaustivity effects, in the sense discussed in Section~\ref{sec:anti-exh-RSA}. 

\subsection{The ``wonky world'' RSA model}
\label{sec:wRSA}

\citet{degen2015wonky}, to our knowledge, is the only proposal within the RSA framework which {is intended to} address {the problem of anti-exhaustivity}, focusing on the case of \textit{some} vs.\ \textit{all}. Suppose that someone throws marbles into a lake. Typically, one expects that if one them sinks, then all of them will. That is, the conditional probability that all marbles sank given that some did is very high. For this reason, in the baseline RSA model, an utterrance of \textit{Some of the marbles sank} (\textit{some} for short)  ends up conveying  that all of the marbles sank.
This is not intuitively a good result. \citeauthor{degen2015wonky} discuss the results of an experiment in which they tested the influence of priors on interpretation, by varying the predicates used in the frame \emph{Some NPs VP}.  Schematically, they collected priors on the number of $NP$s expected to have the property denoted by the $VP$ before any message is presented, and then posteriors after a \textit{some}-message is presented.  While they found an influence of priors on interpretation, {the best-fitting baseline RSA model does not fit the data well} because it assigns too high a probability to the possibility that all the marbles sank after the \textit{some}-message is processed. This was so even though in their implementation, \citeauthor{degen2015wonky} did not include the message \utt{some but not all}, so that the baseline RSA model could not, properly speaking, give rise to genuine anti-exhaustivity (where the listener would increase the probability {that all the marbles sank} after hearing \utt{some} beyond what the priors would warrant). If we add to the set of messages the \textit{some but not all} message and assign it a cost higher than that of other messages but not infinite, the predictions are even worse.

\cite{degen2015wonky} propose a modified RSA model which they argue is much better at predicting the interpretation of \textit{some}-sentences when priors are biased. The core idea is that the listener is uncertain about which priors the speaker is assuming, and she jointly reasons about these priors as well as the state of the world. In a nutshell, when the priors are very biased in favor of \utt{all}, the message \utt{some} has a {relatively} low probability of use. Upon receiving \utt{some}, the listener then reasons that the speaker must have been assuming a less biased prior distribution, which prevents her from deriving anti-exhaustivity. 

In \citeauthor{degen2015wonky}'s model, the listener and the speaker are communicating about some dimension of interests (for instance whether none, some but not all, or all of the marbles sank), but the listener is uncertain about the `background assumptions' that the speaker takes to be common knowledge. For instance, the `default' background assumption might be that all marbles are made out of the same material, so that either all will sink or none will. The `wonky' background assumption would be one where the marbles are assumed not to be made of the same material, so that there is a very high probability that some but not all of them will sink. There are two  values for the background assumption, call them $b_1$ and $b_2$, and the prior probability of a given world $w$ relative to a certain background assumption $b$ is $P(w|b)$.
The listener herself has a prior over $b_1$ and $b_2$ (that is, her prior distribution is a joint distribution over $(w,b)$). The situation described by the model is one where the speaker is (i) fully informed both about $b$ and $w$, and (ii) believes (possibly wrongly) that the value of $b$ is common knowledge. At the same time, the listener---while ignorant about $b$---knows that the speaker knows the value of $b$ and that she believes (wrongly) that the listener knows it.  $P(w|b)$ thus represents both `the probability of $w$ given $b$' and `the probability that $w$ is the case given that the speaker believes that $b$ is common knowledge' (since the speaker believes that $b$ is common knowledge if and only if $b$ is in fact the case).
The model is based on the following equations:

\begin{enumerate}

\item $L_0(w|u,b) \propto P(w|b) \times \sem{u}^{w}$

\item $U_1(u|w,b) = log(L_0(w|u,b)) - c(u)$

\item $S_1(u|w,b) \propto e^{\lambda U_1(u|w,b)}$

\item$L_1(w,b|u) \propto S_1(u|w,b)\times P(w|b)\times P(b)$

\end{enumerate}

The equation for $L_0$ describes the behavior of a listener who believes that the background assumption $b$ is true. The equations for $U_1$ and $S_1$ make sense for a speaker who believes that the listener is assuming the background assumption $b$. The level-1 listener computes the joint posterior distribution on $(w,b)$ using Bayes' rule. This equation captures the following kind of reasoning: given that the speaker used message $u$, I should give higher credence to pairs $(w,b)$ such that the speaker was particularly likely to use $u$ if $b$ and $w$ were in fact the case and the speaker assumed that I know $b$.  

In their experiments, \citeauthor{degen2015wonky} collected priors from participants for different lexical choices, so as to vary the priors, as well as posterior beliefs about, e.g., how many marbles sank given that the message \utt{some of the marbles sank} was used. They evaluated their model by fitting various parameters to the observed data. Among the parameters that are fitted is the prior on background assumptions $b_1$ and $b_2$, where $b_1$ is an assumption that corresponds to a uniform prior (the `wonky world') over $w$, and $b_2$ corresponds to the measured prior over $w$ ($w$ being the number of marbles that sank). They obtained the best fit with a nearly equiprobable prior on $b_1$ and $b_2$, and the model is much better at predicting the observed interpretation that the baseline RSA model.

{However, on the assumption that the measured distribution over worlds truly represents prior probabilities for the relevant cognitive computation, the implementation of the model proposed in \citeauthor{degen2015wonky} is, in a fundamental sense, non-Bayesian. The reason is that this implementation uses the measured distribution as the value for the ``non-wonky''  prior $P(w|b=b_2)$, yet the prior over $w$ that the model actually assigns to the pragmatic listener is the \emph{unconditional} prior $P(w) = P(b_1)\times P(w|b_1) + P(b_2)\times P(w|b_2)$. Because, in the best-fitting model, $b_1$ (the `wonky-world' prior) reaches 0.5, this means that the model attributes to the listener a prior over $w$ which is quite different from the one that is measured experimentally.} {Importantly, what makes this implementation non-Bayesian is the way the data and the model are linked, not the model itself and its equations.}

{This non-Bayesian aspect is not by itself problematic.} The idea would be that, in most situations, including most experimental settings, listeners are simply oblivious to some possibilities which, upon reflection, they might realize have a significant probability.  For instance, they would completely ignore the possibility that the marbles do not have all the same density, might be made of unusual material, etc. In other words, listeners might be unable to access their true prior, and report a prior restricted to a sub-part of the logical universe where things are `normal' enough. However, this does not strike us as very plausible if in fact the `unnatural' background assumption (i.e.\ the abnormal part of the logical universe) has a prior probability as high as 0.5, which is needed for the model to deliver good predictions---while at the same time listeners would entirely ignore this possibility when reporting their priors.
{Alternatively, the parameter being fitted can be thought of as a backoff parameter, or the degree to which prior knowledge should be abandoned in favor of uniform belief. Indeed, w}ith $P(b_1) = 0$, \pocite{degen2015wonky} model just becomes the baseline model based on the measured prior, while with $P(b_1) = 1$, what we get is the baseline model with complete ignorance of the measured prior, and uniform priors instead. In a sense, the proposed model could {therefore} be viewed as {gradually} incorporating the possibility of `base-rate neglect', whereby listeners would not take into account their prior \citep{kahneman1973psychology}.

One possible way of modifying \citeauthor{degen2015wonky}'s model is to drop the assumption that the speaker's belief about `background' assumptions is viewed by the listener as necessarily true. In particular, {such a model would} no longer equate $P(w|$`the speaker believes $b$'$)$ with $P(w|b)$. Rather, {the model now assumes} that the listener uses her own prior over $w$ (which corresponds to the {non-`wonky' priors,} $b_2$) to update her beliefs, and has otherwise a prior over the various background assumptions that the speaker might be entertaining (possibly wrongly). The modified model is described by the following equations:

\begin{enumerate}

\item $L_0(s|u,b) \propto P(s|b) \times \sem{u}^{s}$

\item $U_1(u|s,b) = log(L_0(s|u,b)) - c(u)$

\item $S_1(u|s,b) \propto e^{\lambda U_1(u|s,b)}$

\item $L_1(s,b|u) \propto P(s|b_2)S_1(u|s,b).P(b)$

\item $L_1(s|u) \propto P(s|b_2)\sum\limits_{b} S_1(u|s,b)P(b)$ \\
\end{enumerate}

The important change is at $L_1$, who now uses the measured prior to determine the prior on $w$ ($P(w|b=b_2)$) but does not assume that the speaker shares her prior, and considers all possibilities that correspond to different speaker beliefs about the prior ({under this model,} $P(b)$ does not denote the prior probability that the listener assigns to the background assumption $b$, but the prior probability assigned by the pragmatic listener to the speaker's beliefs about the literal listener's prior). Although arguably more complex, such a model is theoretically advantageous insofar as it allows us to use the measured prior to model the data while still viewing the listener as a fully Bayesian agent. It is bound to predict a stronger influence of the `normal' prior on interpretation than the original model, and as shown in Appendix~\ref{sec:wRSA-antiexh-proof}, it may require higher values of $P(b_1)$ in order to get rid of anti-exhaustivity.

Finally, we point out that besides the prior $P(b_1)$, the wonky/backoff prior $P(w|b_1)$ itself is not uniquely defined and should be seen as a hidden parameter, the value of which substantially affects the output in both versions of the model. Indeed, worlds can be arbitrarily fine-grained, making it impossible to properly define a uniform distribution over $w$. The wonky prior thus has to be a uniform distribution over the cells of some Question Under Discussion (QUD), i.e.\ sets of worlds which only differ along dimensions which are irrelevant to the current discussion (see next section for a formal introduction of the concept).\footnote{To model their data, \cite{degen2015wonky} implicitly assume that the QUD is ``How many marbles sank?'', which is the question participants had to answer, but not necessarily the question they thought the speaker of target utterances was addressing. Under this how-many QUD, the probability assigned by wonky prior to the `all'-situation given that \utt{some} is true is $\frac{1}{15}$. Yet this probability could rise as high as $\frac12$ if the QUD were ``Did all of the marbles sink?'' or fall as low as $2^{-15}$ with ``Which marbles sank?''. Our results suggest that a choice of QUD which increases the probability $P(w_{all}|b_1)$ requires increasing the wonkiness prior $P(b_1)$ accordingly to prevent the model from predicting anti-exhaustivity.} Overall, the degrees of freedom corresponding to the prior on wonkiness and the implicit QUD encoded in the wonky prior raise the question whether the wRSA model offers more than a baseline RSA model with priors that would not be those that are actually measured, but would be fitted so as to get a maximally good correspondence with the data.

In the next sections, we introduce three different RSA models in which sentences can be ambiguous in terms of their literal meaning. This will allow us to compare models which do or do not allow for the presence of exhaustivity operators in the logical forms of sentences (i.e.\ where a sentence could be ambiguous between its `plain' meaning and its exhaustive interpretation). 

\subsection{Supervaluationist RSA}
\label{sec:spector}

\cite{spector2017homogeneity} proposed a model with a completely different purpose in mind, but which can be applied to the case at hand. The model was designed to capture the interpretation of plural definite descriptions, and assumed that their meaning is underdetermined between an existential meaning and a universal meaning. To apply this model to our case, we will  treat sentences as ambiguous between their `plain' meaning and an `exhaustified' meaning. In \pocite{spector2017homogeneity} model, the first-level pragmatic speaker takes into account the fact that sentences may have mutliple meanings:\footnote{This is different from what happens in the Lexical Uncertainty model presented in section~\ref{sec:RSA-LU}, where semantic underspecification is taken into account by the first-level pragmatic listener, but not by the first-level pragmatic speaker.} the speaker is uncertain about which meaning the literal listener will assume, and so chooses her message so as to maximize the \emph{expected utility} of her utterance, i.e.\ the weighed average of the utilities that would be achieved for each possible interpretation, given some prior distribution over interpretations. On top of that, the model incorporates uncertainty about which components of the world state are relevant to the local communicative goal, as formalized with the notion of Questions Under Discussion (QUDs, cf.  \citealt{roberts2012information}). The model views the speaker as wanting to communicate both the answer to a QUD, and the identity of the QUD she is addressing.\\
The model is described by the following equations:

\begin{enumerate}
	\item Ingredients: a set of messages, each of which has a cost, a set of QUDs {(which are denoted with variable $Q$)}, a prior probability distributions over QUDs, interpretation functions ($i$) and world states ($w$), where each of these variables are probalistically independent. $\sem{u}^i(w)$ denotes the truth value of $u$ in $w$ under interpretation $i$.
	\item A QUD $Q$ is viewed as an equivalence relation over worlds, notated $\sim_Q$, and $Q(w)$ denotes the set of worlds equivalent to $w$ relative to $Q$ (i.e.\ worlds which do not differ from $w$ along any dimension relevant to the question denoted by $Q$)
	\item $L_0(w,Q|u, i) \propto P(w)P(Q)\sem{u}^i(w)$
	\item $U_1(u|w,Q) = \sum\limits_{i}P(i) \log(L_0(Q(w),Q)|u,i)) =\sum\limits_{i}P(i)\log(\sum\limits_{v \sim_Q w}L_0(v,Q|u,i)) - c(u)$
	\item $S_1(u|w,Q) \propto \exp(\lambda U_1(u|w,Q))$
	\item For $n \geq 1$, $L_n(w,Q|u) \propto P(w)P(Q)S_n(u|w,Q) - c(u)$
	\item For $n \geq 1$, $U_{n+1}(w,Q|u) = \log(L_n(Q(w),Q|u)) - c(u) = \log(\sum\limits_{v \sim_Q w}\log(L_n(v,Q|u)) - c(u)$
	\item For $n \geq 1$, $S_{n+1}(w,Q|u) \propto \exp(\lambda U_{n+1}(u|w,Q))$
\end{enumerate}

This model can be adapted to our case in the following manner. We now view exhaustification as a \emph{semantic} phenomenon {that involves an operator $exh$} (along the lines of \citealp{chierchia2012scalar}). The message \utt{A} is viewed as ambiguous between $exh(A)$ (= $A \wedge \neg B$) and {its literal interpretation} \utt{A}. The model includes two worlds, $w_a$, $w_{ab}$, and three messages, \utt{A}, $A \wedge \neg B$, $A \wedge B$, where the last two messages have a positive cost. There are two interpretation functions, $i_{lit}$ and $i_{exh}$, which only differ regarding the meaning of \utt{A}: under $i_{lit}$, \utt{A} is true in both $w_a$ and $w_{ab}$, but under $i_{exh}$, \utt{A} is interpreted as $exh(A)$ and true only in $w_a$. Finally, given the two worlds we consider, there are only two possible QUDs: the total question $Q_{total}$ where the speaker wants to communicate the actual world (i.e.\ for either world $w_a$ and $w_{ab}$, $Q_{total}(w) =\{w\}$), and {the partial QUD }$Q_A$ where the speaker only cares about communicating the truth-value of \utt{A}. In this scenario, given that \utt{A} is true in both worlds, this amounts to a question whose answer is already known (i.e.\ for either world, $Q_{A}(w) =\{w_a,w_{ab}\}$). 

Consider now a first-level pragmatic speaker who wants to answer the question $Q_{total}$ and believes $w_{ab}$. {Under the basic assumption that $P(i_{exh})>0$}, the message \utt{A} is not usable at all {for such a speaker}. This is because under $i_{exh}$, \utt{A} is false in $w_{ab}$, and for this reason the utility achieved by \utt{A} under $i_{exh}$ is infinitely negative, hence the expected utility of \utt{A} is itself infinitely negative.\footnote{One of the terms in the sum that defines $U_1(A, |w_a, Q_{total})$, namely $P(i_{exh})\log(L_0(Q_{total}(w_{ab}), Q_{total}|A, i_{exh}))$, is infinitely negative because $L_0(w_{ab},Q_{total}|A, i_{exh}) = 0$.} 
Therefore a speaker who \emph{wants} to convey $w_{ab}$ has to say $A \wedge B$, and so there can't be any anti-exhaustive use of \utt{A}.\footnote{This means that in this model, relative to a maximally fine-grained QUD, a message cannot be used unless it is true on all its possible meanings, and will therefore be interpreted as entailing the conjunction of all its possible meanings. For this reason we call this model the \emph{supervaluationist} RSA model. In philosophical logic, supervaluationism (\citealt{van1966singular,fine1975vagueness}) refers to a trivalent semantics for semantically underspecified sentences (e.g., vague sentences)  where sentences are said to be \emph{supertrue} if they are true under all their possible construals, \emph{superfalse} if they are false under all their possible construals, and undefined otherwise.} Now, if the speaker wants to address $Q_{A}$, then \utt{A} is a good message  in $w_{ab}$, even under $i_{exh}$: even though the exhaustive reading of \utt{A} is false, it still singles out the true answer to the question. As a consequence, from the first-level pragmatic listener's ($L_1$) perspective, the message \utt{A} will still be compatible with the world $w_{ab}$, since $L_1$ cannot categorically exclude that the intended question is $Q_{A}$, and so some influence of the prior probabilities on worlds is expected.
What is important, however, is that there can not be any anti-exhaustivity effect on the interpretation side: under $Q_{total}$, \utt{A} singles out $w_a$, and under $Q_{A}$, it does not {distinguish $w_{ab}$ from $w_a$}, so $L_1$, who is uncertain about the QUD, is bound to increase the probability of $w_a$ compared to that of $w_{ab}$ (because of the possibility that the QUD is $Q_{total}$). The extent to which the interpretation will be exhaustive, i.e.\ to which the posterior probability of $w_a$ will exceed its prior probability, will depend on the priors. If the priors are such that  that $w_a$ is antecedently extremely unlikely, then $L_1$ will think, upon hearing \utt{A}, that probably the speaker meant to address $Q_A$, and so the exhaustivity effect will be small, but still non-zero. For a formal proof, see Appendix~\ref{sec:Spector-antiexh-proof}.

Finally, will a speaker who wants to convey $w_a$ choose \utt{A}  or $A \wedge \neg B$? Suppose the level-1 speaker $S_1$ wants to answer $Q_{total}$ and to communicate $w_a$. Both \utt{A} and $A \wedge \neg B$ are usable. In terms of informativity, $A \wedge \neg B$ is better than \utt{A}: $A \wedge \neg B$ singles out $w_a$ under both $i_{lit}$ and $i_{exh}$, while \utt{A} does so only under $i_{exh}$. But \utt{A} is of course better in terms of cost. The outcome of the tradeoff between informativity and cost will, as in the baseline RSA model, depend on the prior distribution over worlds. If the priors are extremely biased in favor of $w_{ab}$, the informativity disadvantage of \utt{A} will be more pronounced than in a less biased situation (because under $i_{lit}$, the listener would assign a high probability to $w_{ab}$ after interpreting \utt{A}). So $S_1$'s probability of choosing $A \wedge \neg B$ will be an increasing function of the prior probability of $w_{ab}$.

While we have discussed what happens at $S_1$ and $L_1$, we should note that things do not substantially change at $S_2$. $S_2$ still chooses her message depending on the QUD, and, when she wants to convey $w_{ab}$ to answer $Q_{total}$, she will have to use the message \utt{A and B}. This is because $S_2$ thinks she talks to $L_1$, and upon receiving \utt{A}, $L_1$ infers that either the QUD is $Q_A$, or the QUD is $Q_{total}$ and the world is $w_a$. So the message \utt{A} cannot be used to communicate the pair $(w_{ab},Q_{total})$. The fact that the speaker cares about communicating the QUD plays a crucial role here. Furthermore, when she wants to communicate $w_a$ (in the context of $Q_{total}$), the speaker's choice between \utt{A} and $A \wedge \neg B$ will still be sensitive to priors. One difference between $S_2$ and $S_1$ is that $S_2$ is talking to a listener who is able to draw inferences about the intended QUD. Since $S_2$ cares about communicating the QUD (and not just the answer to the QUD), she will take into account $L_1$'s inferences about QUDs. In particular, when the priors are extremely biased in favor of $w_{ab}$, the message \utt{A} prompts $L_1$ to increase the probability she assigns to $Q_A$ (because under $Q_{total}$, \utt{A} was unlikely to be used, since it would communicate the unlikely world $w_a$). This will create an extra incentive for $S_2$ to use $A \wedge \neg B$ when she wants to convey the pair $(w_a,Q_{total})$.\footnote{\label{fn:Spector-conjunction}However, in cases where the prior is, on the contrary, extremely biased in favor of $w_a$, something quite counterintuitive could happen regarding the interpretation of $A \wedge B$: if $w_{ab}$ is now extremely unlikely, $L_1$ will reason that the intended QUD is probably $Q_A$ (since under $Q_{total}$, the message $A \wedge B$ would probably be false), and then interpret $A \wedge B$ as compatible with $w_a$. Furthermore, using $A \wedge B$ could be then an excellent way for $S_2$ to signal that the intended QUD is $Q_A$, and so $A \wedge B$ might become the preferred way of communicating that the intended QUD is $Q_A$, irrespective of whether \utt{B} is true. Note however that for $S_1$, \utt{A and B} is never the best message to convey $(w_a,Q_A)$, so this puzzling behavior disappears as the rationality parameter increases.} 

To conclude, this model categorically excludes the possibility of an anti-exhaustive use or interpretation of \utt{A}, but it still predicts an effect of priors on message choice and also on interpretation, in the sense that the exhaustivity inference triggered by \utt{A} will be modulated by priors, as will be the speaker's choice between the two messages \utt{A} and $A \wedge \neg B$. This model essentially makes the prediction that \utt{A} triggers the inference that either the speaker considers \utt{B} to be irrelevant, or the speaker believes \utt{B} to be false. It is thus in line with some recent proposals which argue that exhaustivity inferences are obligatory when relevant, such as \cite{buccola2019obligatory}.

\subsection{The Lexical Uncertainty Model}
\label{sec:RSA-LU}

The Lexical Uncertain Model (LU for short, see \citealp{bergen2016pragmatic}) is another model in which messages can be ambiguous. It includes all the ingredients of the baseline RSA model, and adds to it a set of interpretation functions. {Just like the model presented in the previous section, t}he literal listener $L_0$ is relativized to a specific interpretation function (hence there are as many literal listeners as there are interpretation functions). The first-level pragmatic speaker $S_1$ is also relativized to a specific interpretation function, insofar as she chooses messages for a literal listener who has a particular interpretation function (this is unlike the supervaluationist model discussed in the previous section, where $S_1$ is uncertain about which literal listener she is talking to). It is only at the level of the first pragmatic listener $L_1$ that uncertainty about meaning is factored in. $L_1$ has a prior distribution over worlds and interpretation functions, which are assumed to be probabilistically independent, and uses Bayes' rule to infer a joint posterior distribution over worlds and interpretation functions. Her posterior distribution over worlds is obtained by marginalizing on interpretation functions. Then higher-level listeners/speakers are defined just as in the baseline model (since $L_1$'s behavior is no longer dependent on a specific interpretation function).  The model is defined by the following equations:

\begin{enumerate}
	\item $L_0(w|u, i) \propto P(w)\sem{u}^i(w)$
	\item $U_1(u|w,i) = \log(L_0(w|u,i)) - c(u)$  
	\item $S_1(u|w,i) \propto \exp(\lambda U_1(u|w,i))$
	\item $L_1(w|u) \propto P(w)\sum\limits_i P(i)S_1(u|w,i)$
	\item For $n \geq 1$, $U_{n+1}(u|w) = \log(L_n(w|u)) - c(u)$
	\item For $n \geq 1$, $S_{n+1}(u|w) \propto \exp(\lambda U_{n+1}(u|w))$
	\item For $n \geq 2$, $L_n(w|u) \propto \log(S_n(w|u))$
\end{enumerate}

We can then consider two versions of this model, depending on the set of interpretation functions we include in the model. In one version, which we call the EXH-LU model, the possible interpretation functions are, like in the previous model, $i_{lit}$ and $i_{exh}$. This is the version of LU used in \cite{franke2020theory}. In the second version, which we call the FREE-LU model, we add to these two interpretation functions a third one, $i_{\textit{anti-exh}}$, where \utt{A} means $A \wedge B$. This version corresponds to the unrestricted LU model used in \cite{bergen2016pragmatic,potts2016embedded}, in which the possible meanings of a message consist of all the possible ways of strengthening their literal meaning. In the EXH-LU model, symmetry between \utt{A and B} and \utt{A and not B} is broken in the grammar, while the FREE-LU, just like the baseline model, can only break symmetry with cost and priors.

In the EXH-LU model, if the two interpretation functions are equiprobable and as long as the cost of $A \wedge B$ does not exceed that of $A \wedge \neg B$, it can be shown that there cannot be any anti-exhaustivity effect at $L_1$. However, anti-exhaustivity effects can arise  at $S_2$ (in the sense that $S_2$ may use \utt{A} to convey $w_{ab}$) and at higher-level listeners. This is essentially because at higher levels of recursion, the model is indistinguishable from the baseline model. If the priors are sufficiently biased in favor of $w_{ab}$, $L_1(w_{ab}|A)$ will itself be close to 1 (though not higher than $P(w_{ab})$), and so $S_2$ might be more likely to use \utt{A} in $w_{ab}$ than in $w_a$ (because \utt{A and not B} would be a much better message that \utt{A} in $w_a$), following exactly the same logic as in our discussion of the baseline RSA model. 

As to the FREE-LU model, unsurprisingly, it does not exclude anti-exhaustive inferences even at $L_1$, because one of its interpretation functions makes \utt{A} equivalent to $A \wedge B$.

\subsection{The Lexical Intentions Model}

Finally, we {will} consider {another} model in which messages can be ambiguous---the Lexical Intentions Model, recently proposed by \cite{franke2020theory}.\footnote{\cite{franke2020theory} also discuss a Global Intentions model, which is equivalent to the LI model for the cases we consider in this manuscript.} In this model each message can again have multiple meanings. The crucial difference with the LU model is that the first-level speaker jointly chooses a message and an interpretation {function} for this message, so as to maximize the conditional probability of the intended world given the meaning that is chosen. This corresponds to a speaker who chooses a message with a certain interpretation in mind, and based on a literal listener who will always interpret the message under the interpretation that the speaker has in mind. If we again consider two possible interpretations {functions}, $i_{lit}$ and $i_{exh}$,\footnote{This amounts to counting unambiguous messages twice each, since they can be paired to both $i_{lit}$ and $i_{exh}$, with no change in meaning. This increases the probability of unambiguous messages to be used, all else being equal, and  is in line with \pocite{franke2020theory} implementation (Michael Franke, p.c.). Other choices are possible, with no important qualitative change.}  then the best message-interpretation pair for a speaker who wants to convey $w_a$ is the pair $(A, i_{exh})$, which is true only in $w_a$ and is less costly than any pair $(A \wedge \neg B, i)$.

The first level listener, $L_1$, performs a joint inference both on worlds and intended interpretation. The prior distribution over $i$ is uniform (and for this reason can be entirely ignored when stating proportionality statements). The model is characterized by the following equations:

\begin{enumerate}
	\item $L_0(w|u,i) \propto P(w)\sem{u}^i(w)$  
	\item $U_1(u,i|w) = \log(L_0(w|u,i)) - c(u)$
	\item $S_1(u,i|w) \propto \exp(\lambda U_1(u,i|w))$
	\item $L_1(w,i|u) \propto P(w)S_1(u,i|w)$	\hfill priors on $i$ are {implicitly} uniform
\end{enumerate}

Predictions about the speaker's choice of a message and the listener's inferences about the world are obtained by marginalizing over meanings:

\begin{enumerate}
\setcounter{enumi}{4}
\item $S_1(u|w) = \sum\limits_i S_1(u,i|w)$
\item $L_1(w|u) = \sum\limits_i L_1(w,i|u) \propto P(w)\sum\limits_i S_1(u,i|w)$
\end{enumerate}

We will consider a model with the same set of worlds and messages as before, and where the message \utt{A} is ambiguous between the plain and the exhaustified meaning, while the two other messages are unambiguous. It can be proved that as long as the cost of $A \wedge B$ is no greater than that of $A \wedge \neg B$, \utt{A} can never be the preferred message to convey $w_{ab}$, and \utt{A} is never interpreted anti-exhaustively. The reason is that $S_1$  will never be more likely to use \utt{A} in $w_{ab}$ than in $w_a$. In $w_a$, \utt{A} is always the best message to choose (under the interpretation $i_{exh}$). In $w_{ab}$, \utt{A} might be the best message to use if the priors are extremely biased in favor of $w_{ab}$, but even in such a case it will be no more likely to be used in $w_{ab}$ than in $w_a$ because (i) it competes against \utt{A$\wedge$B} which is less costly that \utt{A$\wedge\neg$B}, and (ii) it must be interpreted according to $i_{lit}$, and face a small but positive informativity penalty compared to the interpretation $i_{exh}$ in $w_a$.
Now, while \cite{franke2020theory} do not define higher-level speakers and listeners for the Lexical Intention model, the most natural way to extend their proposal to higher-level speakers and listeners would be to define them as in the baseline RSA model, as shown below, in which case anti-exhaustivity effects could reappear at higher levels of recursion, as discussed in the previous section in relation with the LU model.\footnote{{Note that since $L_1$ infers the interpretation $i$ intended by $S_1$  based on the surface form of the message only, there is no way for $S_2$ to influence the interpretation inferred by $L_1$ for a given message. Therefore, the intention mechanism cannot be recursively extended beyond $S_1$.}}

\begin{enumerate}
\setcounter{enumi}{6}
\item For $n \geq 1$, $U_{n+1}(u|w) = \log(L_n(w|u)) - c(u)$
\item For $n \geq 1$, $S_{
n+1}(u|w) \propto \exp(\lambda U_{n+1}(u|w))$
\item For $n \geq 2$, $L_n(w|u) \propto P(w)S_n(u|w)$
\end{enumerate}

\section{Experiment}
\label{sec:experiment}

The experiment was pre-registered on OSF (\url{https://osf.io/7xts3/}). All data and scripts are available at\\ \url{https://github.com/Alex-Cremers/RSA-Exh-Priors}.

\subsection{Background and motivation}
\label{sec:empirical-background}
Several experimental studies have investigated the influence of priors on exhaustive interpretations, starting with \cite{degen2015wonky}. In the analogous case of \utt{some}/\utt{all}, they found that the prior probability for the all-situation only had a small effect on the interpretation of \utt{some}, which was overwhelmingly interpreted as \utt{some but not all}, and this was the motivation for the wRSA model.
\cite{javangula2019} investigated effects of priors and QUD manipulations on comprehension, and found  that priors modulated the strength of exhaustivity inferences, but did not find evidence for anti-exhaustivity.
\cite{schreiber2021narrow} looked directly at the type of exhaustivity inferences discussed in the introduction, and found that exhaustive interpretations are essentially constant across domain sizes and priors. Finally, \cite{wilcox2019} looked at the effect of priors on both production and comprehension. They only found a small effect on exhaustive interpretations, and did not observe the predicted effect on production, but the results were too noisy to draw firm conclusions.

The goal of our experiment was to improve on previous work in several respects. First, we wanted to obtain cleaner results on production, where the predictions of the {baseline} RSA are intuitively correct, at least qualitatively. Second, none of the cited studies measured priors from the same participants who took part in the comprehension/production task. We decided to gather prior data from the same participants we tested for comprehension/production in order to maximize our chances to find correlations between priors and linguistic behavior.\footnote{Using a noisy predictor leads to systematic underestimation of the coefficient in a linear regression (a phenomenon known as \emph{regression dilution}), so reducing the noise on the prior measure should increase the measured correlation with comprehension/production.}

\subsection{Methods and materials}

\subsubsection{Experimental design}

\begin{figure}
\centering
\framebox{
	\includegraphics[width=.9\textwidth]{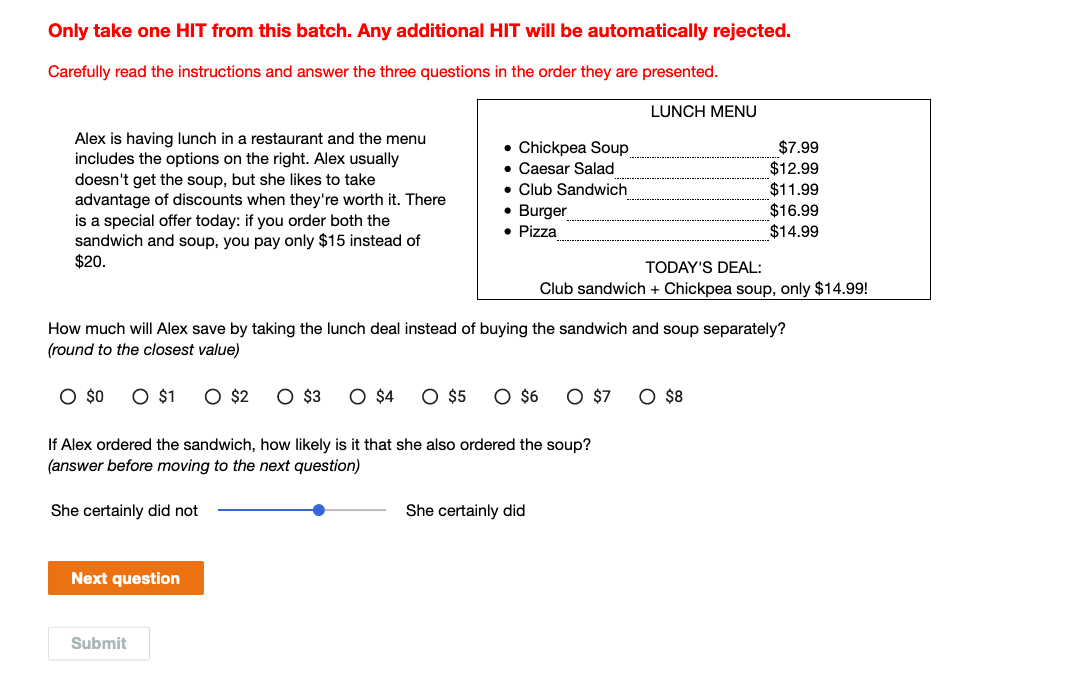}
}
\caption{Example survey, before submitting answer to the prior question.}
\label{fig:example_survey}
\end{figure}

Participants were recruited on Amazon's Mechanical Turk, and the survey was hosted directly there. We first introduced a context in which a character, Alex, is choosing what she will have for lunch. An example is given in Fig.~\ref{fig:example_survey}. Among the options are a sandwich and a soup, which can be bought together for a discount. The survey focuses on whether Alex will buy only the sandwich (our $w_a$ situation), or both the sandwich and the soup (a $w_{ab}$ situation). We manipulated the price of the lunch deal from \$12 (the soup comes for free with the sandwich) to \$19 (only \$1 discount), in order to generate various expectations about Alex's preference between the lunch deal and the sandwich alone.

We first asked participants to calculate the value of the discount (i.e.\ to subtract the last two prices in the context story). This was an attention check allowing us to filter out participants who did not pay enough attention, but it also drew attention to our manipulation in order to maximize its effect. We then measured participants' prior belief in Alex ordering the soup conditional on her ordering the sandwich (i.e.\ $P(w_{ab}|\{w_a,w_{ab}\})$), using a slider. After answering the question prior, participants could click a ``next question'' button which brought them to either a production question or a comprehension question, and froze the slider for the prior question (so they could still see their response but could not edit it).

In the production survey, the last question was \Next, and participants had to type in their answer (up to 50 characters). The rationale {for the free response} was to make it more effortful to provide longer messages, thereby increasing the likelihood that cost would factor in participants' choice of a message (in contrast with a multiple choice question where choosing the most informative message has no disadvantage). {Free response} also allowed us to record the length of the different messages, which we {used} as a proxy for cost. We manipulated the information given about Alex's order so as to place participants in either a $w_a$ situation or a $w_{ab}$ situation.

\ex. \label{ex:production_stimuli}
\begin{minipage}[t]{.7\textwidth}
In fact, Alex ordered \{ only the sandwich / both the sandwich and the soup\}.\\
What do you think she would answer to someone who asks what she ordered?\\
\underline{{\color{gray} \texttt{\dots  text response field \dots}} \phantom{p}\rule{15em}{0pt}}
\end{minipage}

In the production survey, the next question was \Next, which participants could answer using a slider identical to the one used for measuring their prior. This question presupposed that Alex did order the sandwich, and measured participants' posterior belief $P(w_{ab}|u)$, where the utterance was either \utt{A} (``the sandwich'') or \utt{A\&B} (``the sandwich and the soup''). Crucially, since their answer to the prior question was still visible, it was very easy for participants to intentionally place the slider lower or higher for the posterior (i.e.\ to indicate an exhaustive or anti-exhaustive interpretation, respectively).

\ex. You ask Alex what she ordered, and she answers \{``The sandwich" / ``The sandwich and the soup''\}.\\
Given her answer, how likely is it that she also ordered the soup?

Production and comprehension data were collected separately to avoid confusion. In all, there were $8\times(2+2)$ versions of the survey: eight possible prices for the lunch deal, two possible worlds for the production question and  two possible utterances for the comprehension question.

\subsubsection{Participants}

Mechanical Turk workers were recruited using the following filters: US Location, HIT approval of at least 98\% {(we initially tried 99\% but too few Turkers met the criterion)}{}, number of approved HITs greater than 1000. Participants were paid {\$0.47} for their participation (the survey took about 3min to complete). 

For each of the 8 prices, data from 20 participants were collected for utterance  \utt{A} (comprehension) and world $w_a$ (production), and from 10 participants for utterance \utt{A\&B} (comprehension) and world $w_{ab}$ (production).\footnote{A pilot study had shown very little variation in participants' behavior in the latter two conditions.} In total, 239 participants were recruited for the comprehension survey and 237 for the production survey (goal: 240 for each: $8 \times (20+10)$). Among them, 222 and 224 passed the comprehension {check} respectively, amounting to an error rate of about 6\%.

\subsubsection{Data analysis}

The production data were coded into the following 6 categories (examples provided): \utt{A} (``the sandwich''), \utt{A\&B} (``the sandwich and the soup'', ``the lunch deal''), \utt{A\&$\neg$B} (``just the sandwich''), \utt{B} (``soup''), \utt{$\neg$B} (``I didn't want the soup''), and Other/NA (``I wasn't hungry today''). The number of characters used to convey the message was recorded. NA responses were excluded from the analyses. Prior measures were compressed from $[0,1]$ to $[.005,.995]$.

We first ran theory-neutral analyses to test the association between the priors and posteriors in comprehension data with linear regression, and the association between priors and production with binomial (logistic) regressions (excluding literally false messages, e.g. \utt{A\&B} in the $w_a$ world). We then fitted the parameters of each of the models of interest to our data set. We only observe 1 \utt{B} response and 12 \utt{$\neg$B} responses, so we decided to merge them with \utt{A\&B} and \utt{A\&$\neg$B} respectively for model fits. For production, strictly-speaking, no linking hypothesis is needed since each model immediately generates a prediction for the likelihood to observe any given message. However, models often predict probabilities of $0$, so following \cite{franke2020theory}, we added an error parameter to our model of production data: if $S(u)$ is the theoretical predicted probability {for observing} message $u$, we used $\frac{S(u)+\epsilon}{1+n\epsilon}$ as our predictor, where $n$ is the number of candidate messages. The parameter $\epsilon$ was fitted together with model parameters. For comprehension, the models return a point estimate for the posterior probability. We assumed that data is normally distributed around the prediction, with censoring at 0 and 1 (the boundaries of the slider), as in tobit regression. This is a direct improvement on the usual least-square fit, which amounts to assuming an uncensored normal distribution (including impossible values beyond 0 and 1). We assumed different noise parameters $\sigma_a$ and $\sigma_{ab}$ for the \utt{A} and \utt{A\&B} condition.

Parameter estimation was done by maximizing the likelihood of the whole dataset (production and comprehension). Models were evaluated by their AIC.

\subsection{Results}

\subsection{Preliminary analyses}

\begin{figure}
\centering
\begin{subfigure}[t]{.475\textwidth}
\centering
\includegraphics[width=\textwidth]{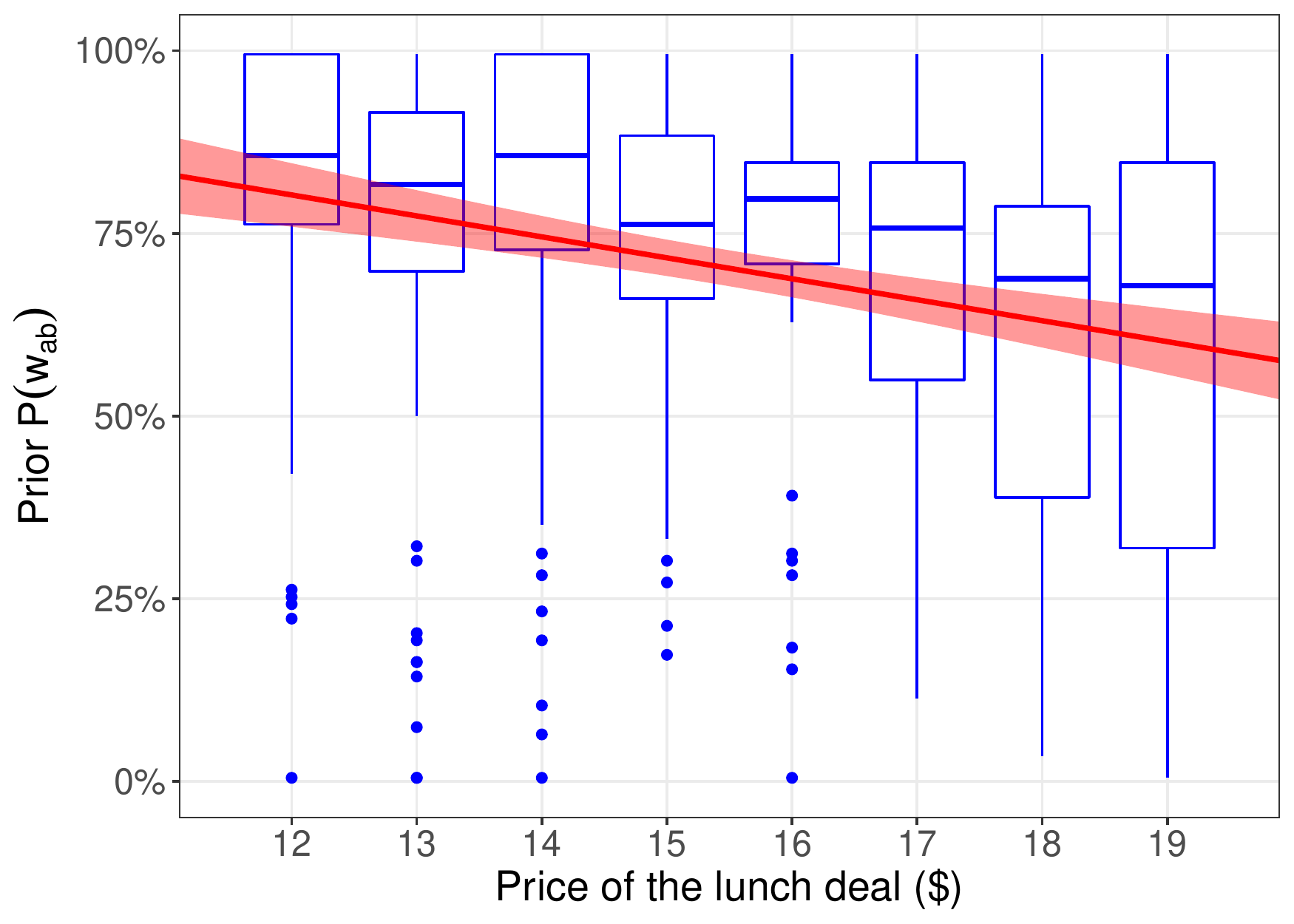}
\caption{Effect of the price manipulation on conditional prior beliefs $P(\text{soup}|\text{sandwich})$, with linear correlation line.}
\label{fig:prior_price}
\end{subfigure}\hfill%
\begin{subfigure}[t]{.475\textwidth}
\centering
\includegraphics[width=\textwidth]{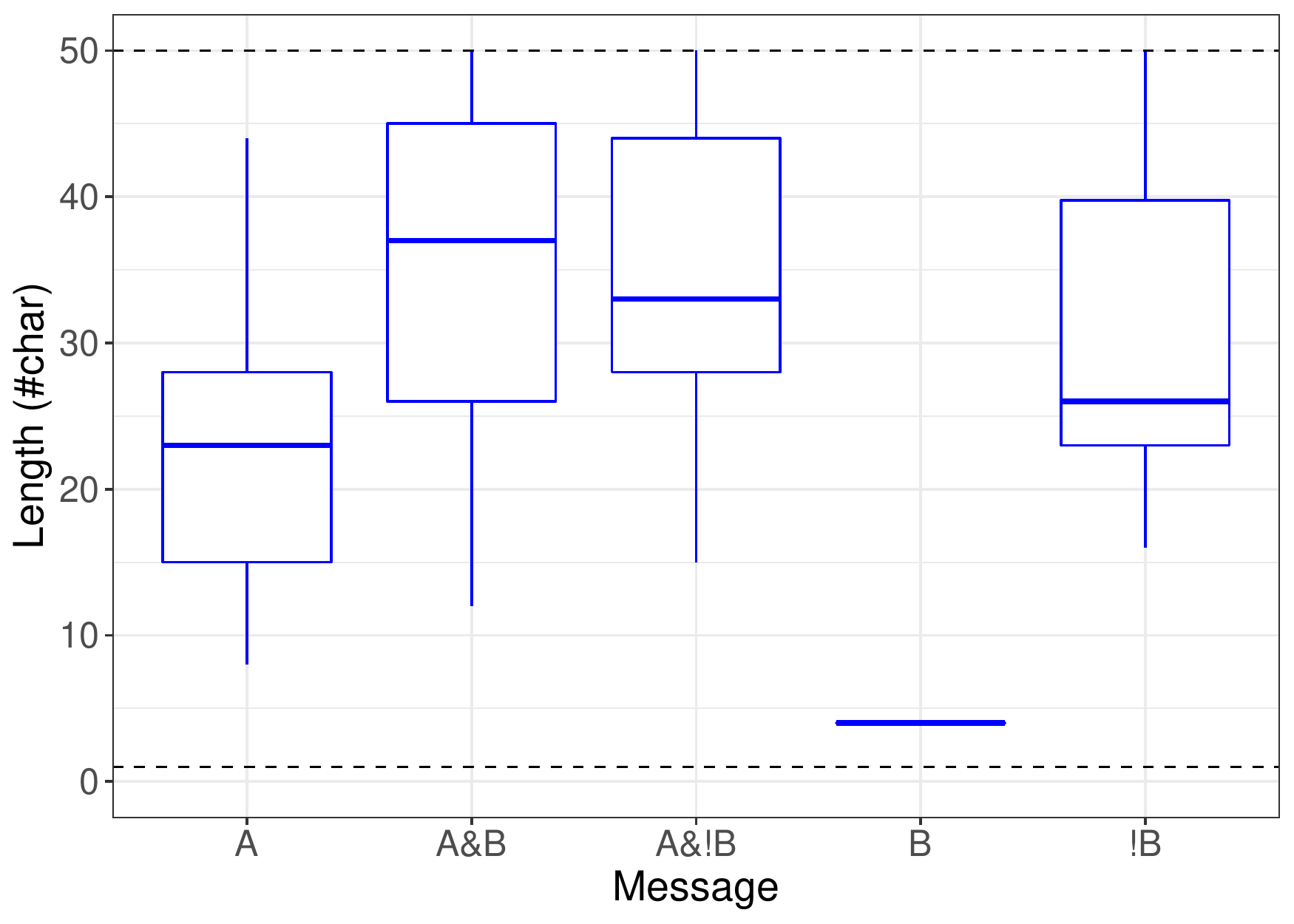}
\caption{Length of elicited responses in number of characters, by expressed message, with lower (1) and upper (50) limits.}
\label{fig:length_message}
\end{subfigure}\\
\begin{subfigure}[t]{.475\textwidth}
\centering
\includegraphics[width=\textwidth]{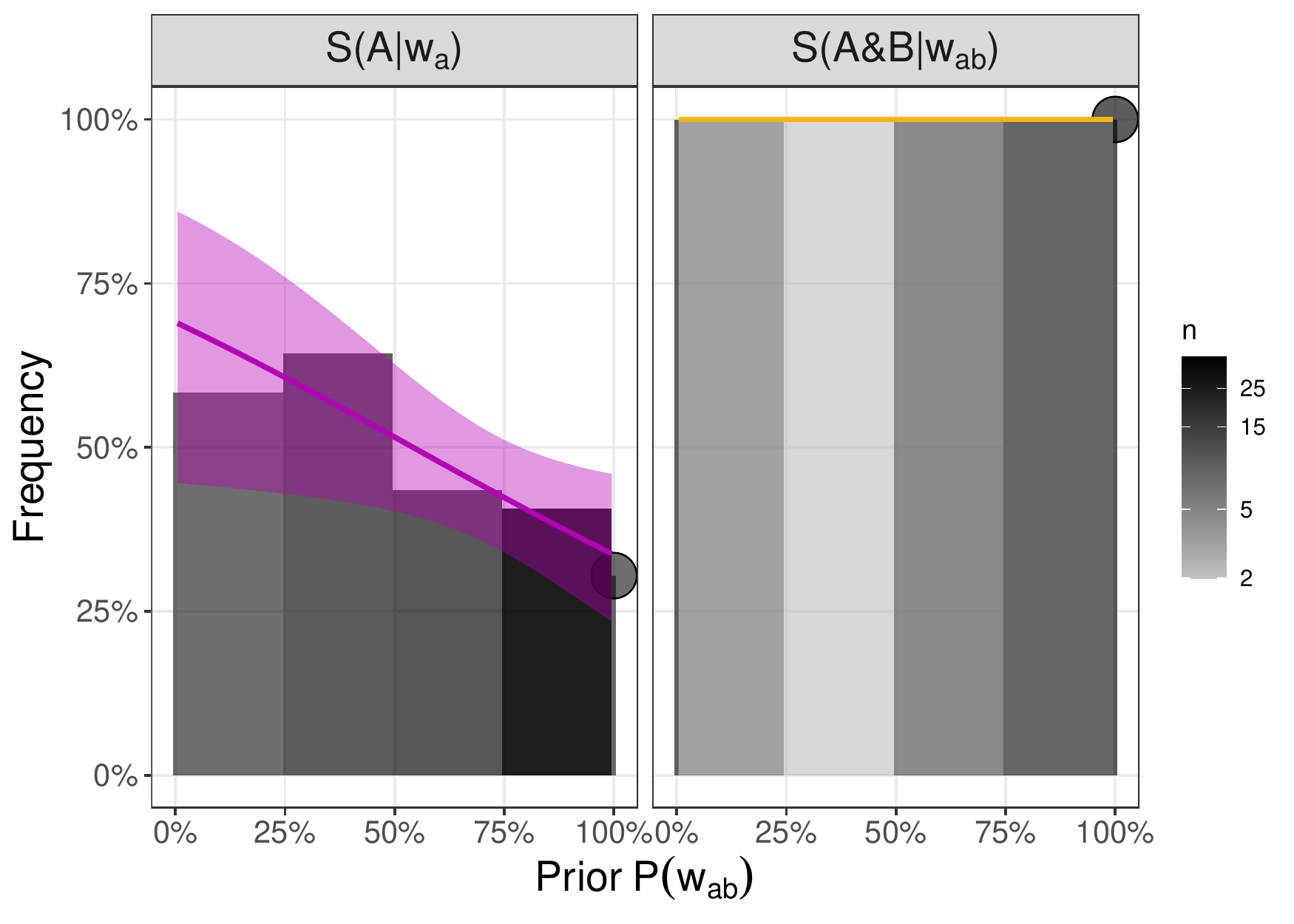}
\caption{Effect of prior beliefs on production, after exclusion of literally false messages. For plotting (and only for plotting), priors are binned into $\left\{[0,24],[25,49],[50,74],[75,99],\{100\}\right\}$, with the last bin represented by a large dot. Shade indicates sample size in each bin. The colored lines correspond to logistic regressions.}
\label{fig:production-prior}
\end{subfigure}\hfill%
\begin{subfigure}[t]{.475\textwidth}
\includegraphics[width=\textwidth]{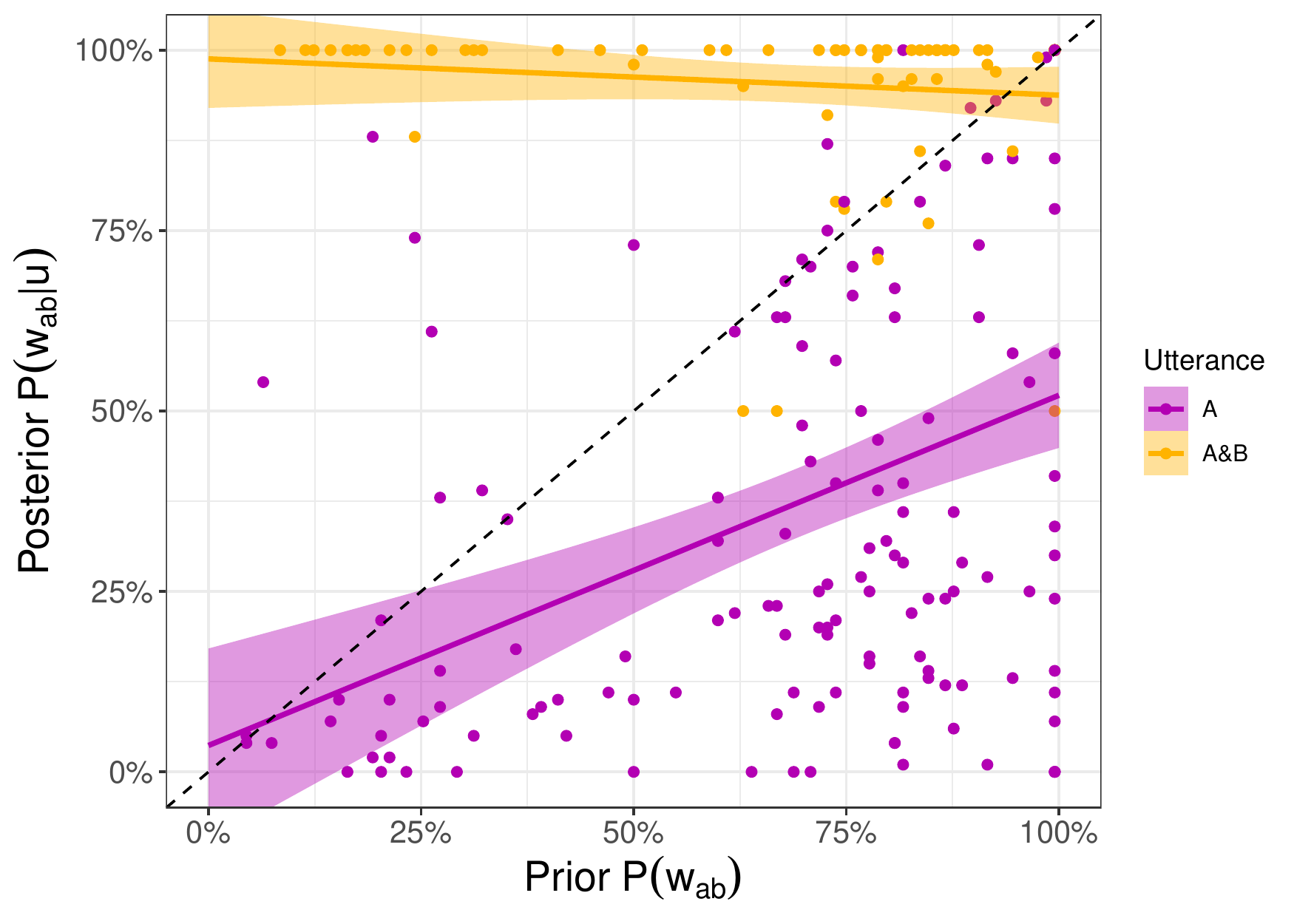}
\caption{Effect of prior beliefs on comprehension for each utterance. The bisecting dashed line indicates the ``posterior=conditional prior line'', which represents a literal interpretation of utterance \utt{A}. Points below this line indicate exhaustivity, while points above the line indicate anti-exhaustivity {(for the A message)}. The colored lines correspond to linear regressions.}
\label{fig:comprehension-prior}
\end{subfigure}
\caption{Descriptive results for prior elicitation, production and comprehension questions.}
\end{figure}

\paragraph{Priors:} The reported priors covered the whole slider range, although they were biased towards higher values ($M=.70, SD=.27$), as intended. The price manipulation had a clear effect on the reported prior ($F(1,409)=33, p<.001$). In all the following, our predictor will be the measured prior for each participant, and we ignore the price variable.

\paragraph{Production:} We observed all categories of responses, but \utt{B} appeared only once as ``SOUP''. The otherwise shortest message on average was \utt{A} ($M=22.8, SD=8.9$), followed by \utt{$\neg$B} ($M=30.5, SD=12$),  then \utt{A\&$\neg$B} ($M=35.7, SD=9.7$, often expressed as ``just the sandwich'') and finally \utt{A\&B} ($M=35.8, SD=11$, expressed either as a conjunction or ``the lunch deal''). We observed a small effect of priors on the production of an exhaustive form to convey $w_a$ ($\chi^2(1) = 4.95, p=.026$): when the prior for the soup is higher, participants are more likely to explicitly say that they didn't order it. {In contrast, \utt{A} was never used to convey $w_{ab}$ (\utt{A\&B} was always used), and so there was no influence of priors regarding the choice of a message in $w_{ab}$.}

\paragraph{Comprehension:} We observed an effect of prior on the interpretation of \utt{A} ($\beta=.49$, $F(1,145) = 28.2, p <.001$), {but no evidence for the kind of anti-exhaustivity predicted by the RSA: 15 out of 147 responses to the target comprehension survey indicated a posterior value higher than the prior, but while the base RSA predicts anti-exhaustivity to occur when the prior is high, we found that the probability to observe such a response was inversely correlated with the prior (logistic regression: $\beta=-.91, z=-3.5, p<.001$). This is exactly what we would expect if these responses were the result of random errors, which are more likely to fall above the prior when the prior is lower.}
By contrast, there was no effect of priors on the interpretation of \utt{A\&B} ($\beta=-.050$, $F(1,73) = 1.2, p=.28$).

\paragraph{Connecting production and comprehension:}

We can already see that participants who took the comprehension survey do not behave as Bayesian listeners with respect to participants' behavior in the production survey. Indeed, no participant produced \utt{A} in the $w_{ab}$ condition (0/72), so a Bayesian listener {making interpretations based on a speaker model that is similar to our measured production data} should conclude that $P(w_{ab}|\utt{A})\approx0$, irrespective of the prior. Yet {comprehension} responses to the \utt{A} condition show a clear effect of prior on comprehension, with posterior values high above 0. This does not mean that listener participants are not Bayesian, but if they are, their model of the speaker does not match what participants do in the production survey.

Conversely, we can fit the costs and rationality parameters of an RSA-speaker addressing a listener who behaves like our participants in the comprehension task.\footnote{The empirical listener was modeled as a linear regression on arcsine-transformed priors and posteriors. We tested polynomial fits up to degree 5, but a linear fit was BIC-optimal.} The resulting RSA speaker yielded a good fit of the data, but with {implausible} parameters $(\lambda=1.6,c_\text{A\&B}=-8.9, c_\text{A\&$\neg$B}=.34)$. In short, the model avoids predicting uses of \utt{A} to convey $w_{ab}$ by assigning a large negative cost to the message \utt{A\&B}. Forcing the model to assign a positive costs to the messages \utt{A\&B} and \utt{A\&$\neg$B} immediately worsens the predictions (log-likelihood falls from $-131$ to $-137$ and the model predicts \utt{A} to be used to convey $w_{ab}$ with probability up to $.19$).

In short, participants in the comprehension task are not Bayesian with respect to {the behavior of} participants in the production task. {Likewise, participants in the production task}---in the sense of the RSA---do not seem to be rational with respect to participants in the comprehension task either.

\paragraph{Preliminary discussion:}

We didn't detect any anti-exhaustivity effect {except for a few data points that would be best explained as random errors}.  In comprehension, the probability assigned to $w_{ab}$ after receiving the message \utt{A} was influenced by prior probabilities but was never higher on average than its prior probability{, except for very low prior values where a few outliers are sufficient to bring the posterior value over the prior}. In production, \utt{A} was never used to convey $w_{ab}$. These results thus provide \emph{prima facie} evidence in favor of models that cannot predict anti-exhaustivity and against those that can. We now turn to evaluating alternatives to the base RSA model.

\subsection{Models evaluation}

\subsubsection{Tested implementations}

We tested how well variants of the baseline RSA model fit our production and comprehension data. Besides the baseline RSA model, we evaluated the following models:
\begin{itemize}
\item The ``wonky world'' RSA model (wRSA) of \cite{degen2015wonky}, as well as the Bayesian variant described in \S\ref{sec:wRSA} (BwRSA). Both versions come with an extra parameter, which corresponds to the ``wonkiness'' prior.
\item Spector's supervaluationist model (svRSA), as described in \S\ref{sec:spector}. Our experiment did not measure participants' assumptions about the QUD, so we fit the prior on the total QUD as a parameter of the model. We test two versions of the model which differ in their definition of the speaker: svRSA1 models production as a mixture of the speakers trying to convey the partial QUD and the total QUD (in proportion of the prior on QUD), while svRSA2 simply models production as a speaker who tries to convey the total QUD. We use a uniform prior on interpretations (exhaustive or literal).
\item The lexical uncertainty model, in its free version (FREE-LU, which allows both exhaustive and anti-exhaustive strengthening{~in line with \citealp{bergen2016pragmatic}}) and in its grammatical version (EXH-LU, which only considers strengthening derived from a grammatical theory of exhaustification{, as in \citealp{franke2020theory}}). For both models, we used a uniform prior on interpretations.
\item The lexical intentions model with grammatical exhaustification. Here again we test two versions of the model, which differ {in terms} of production: RSA-LI1 uses the marginal $S_1$ speaker, while RSA-LI2 uses $S_2$. The crucial difference is that $S_1$ picks simultaneously a message and a meaning in order to communicate a world, while $S_2$ simply picks a message and talks to a listener who has already factored in ambiguity.
\end{itemize}

All models use $L_1$ to model comprehension, and all models except RSA-LI1 use the $S_2$ speaker to model production. This is because in the wRSA and the LU models, $S_1$ is a theoretical construct that is not readily interpretable, as it is relativized to a lifted variable. Furthermore, the choice of $L_1$ rather than $L_2$ was motivated by the observation that the comprehension participants cannot be Bayesian with respect to the empirical speaker. The prediction of all models were computed under the assumption that the possible utterances are \utt{A}, \utt{A\&B} and \utt{A\&$\neg$B} and the worlds to convey are $w_a$ and $w_{ab}$. Models are parametrized by the rationality parameter $\lambda$, the costs of \utt{A\&B} and \utt{A\&$\neg$B} relative to \utt{A} ($\Delta_{ab}$ and $\Delta_{a\neg b}$), and an extra prior parameter for the wonky RSA and the two supervaluationist models. The first three parameters are constrained to be positive, and the prior parameters are constrained to $[0,1]$. Appendix~\ref{app:implementations} gives the exact implementation for each model. For additional details, see the R script in Supplementary Materials.

\subsubsection{Results}

\begin{figure}[p]
\centering
\begin{subfigure}{.95\textwidth}
\centering
\includegraphics[width=\textwidth]{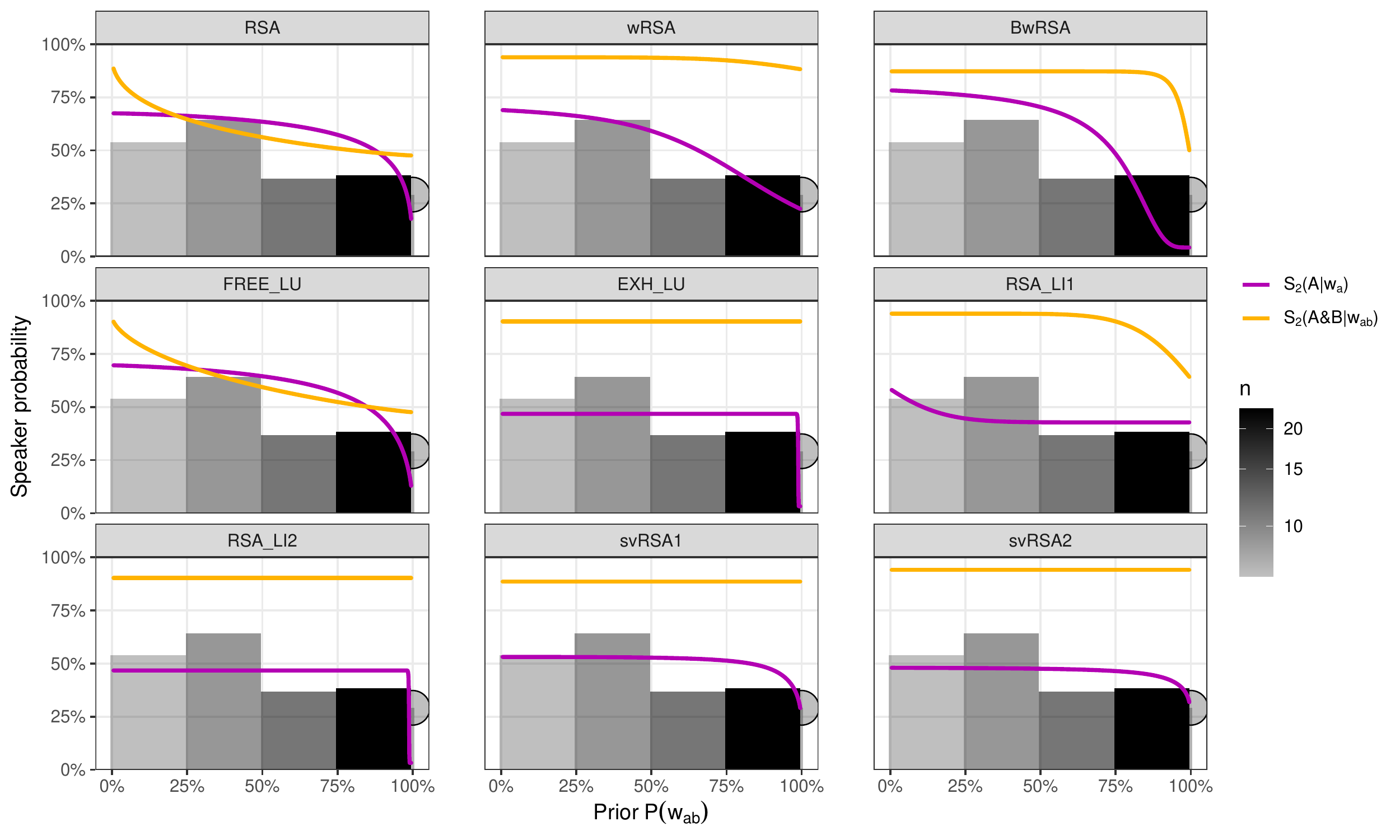}
\caption{Model fits on the production data. We only present the data from the $w_a$ world as the production for the $w_{ab}$ world consisted only of \utt{A\&B} (i.e., the orange line should be as high as possible). Note that unlike Fig~\ref{fig:production-prior}, this plot includes literally false messages (captured by the noise parameter in the model fit but discarded in the preliminary analysis).}
\end{subfigure}\\
\begin{subfigure}{.95\textwidth}
\centering
\includegraphics[width=\textwidth]{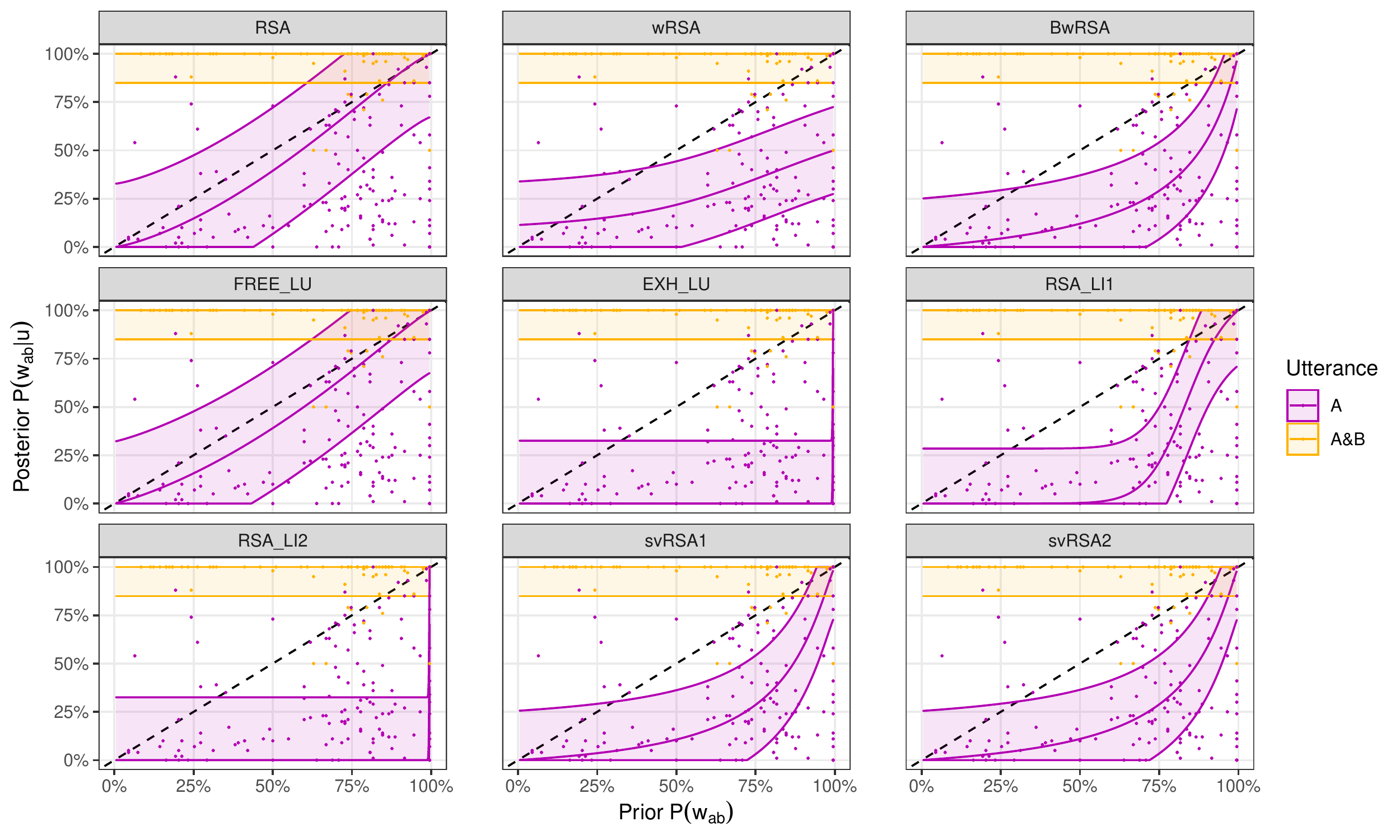}
\caption{Model fits on the comprehension data. We represent the median and quartile predictions (given the censored gaussian model for noise, the theoretical predictions correspond to the median, not the mean).}
\end{subfigure}
\caption{Model fits on production and comprehension. Likelihood was maximized on the combined data.}
\label{fig:free_costs_fit}
\end{figure}

Figure~\ref{fig:free_costs_fit} shows the model fits against the data, and Table~\ref{tab:free_costs_table} gives the parameter estimates. The best models are  the wonky world RSA model (wRSA) and the first version of supervaluationist model (svRSA1), which fit the data equally well (wRSA is better at capturing comprehension and svRSA1 better at capturing production). The fit of comprehension data illustrates our earlier point about how the wRSA model is, in some sense, non-Bayesian, since its predictions are not constrained to 0 and 1 when the prior is 0 and 1, respectively. Note as well that RSA-LI2 and EXH-LU are strictly equivalent when the priors on interpretation functions are uniform and the costs are null, which happens to be where the maximum likelihood is reached.

Interestingly, all models but the supervaluationist model (svRSA) end up assigning a cost of 0 to \utt{A\&B}---presumably to minimize anti-exhaustivity---, while the svRSA model assigns it a very high cost (presumably so that \utt{A\&B} is never used to convey the partial QUD, which would allow its use in the $w_a$ world). Since \utt{A\&B} and \utt{A\&$\neg$B} had roughly the same length in participants' responses (35.8 vs.\ 35.7 characters on average), we refitted the models with an added constraint $\Delta_{ab}=\Delta_{a\neg b}$ as a post-hoc analysis. From a theoretical point of view, this analysis allows us to evaluate how much each model relies on costs to break symmetry. 

\begin{table}[ht]
\centering
\begin{tabular}{|c|c|c|c|c|c|l|l|c|}
  \hline
\multicolumn{1}{|c|}{Model} & \multicolumn{1}{c|}{$\lambda$} & \multicolumn{1}{c|}{$\Delta_{ab}$} & \multicolumn{1}{c|}{$\Delta_{a\neg b}$} & \multicolumn{1}{c|}{$\xi$}& \multicolumn{1}{c|}{$\sigma_a$} & \multicolumn{1}{c|}{$\sigma_{ab}$} & \multicolumn{1}{c|}{$\epsilon$}  & \multicolumn{1}{c|}{AIC} \\ 
  \hline
wRSA & 3.9 & 0 & .37 & .86 & .33 & .22 & .022   & 492 \\ 
  svRSA1 & .68 &  44 & .20 & .89 & .38 & .22 & .003  & 492  \\ 
  svRSA2 & .54 & 131 & 0 & .89 & .38 & .22 & .022& 509  \\ 
  RSA-LI1 & 8.6 & 0 & .054 &  & .42 & .22 & .022 & 551  \\
  BwRSA & 5.8 & 0 & .36 & 1 & .37 & .22 & .052 & 561  \\ 
  EXH-LU & 1.0e+3 & 0 & 0 &  & .48 & .22 & .037 & 593 \\ 
  RSA-LI2 & 1.0e+3 & 0 & 0 &  & .48 & .22 & .037  & 593 \\ 
  FREE-LU & .57 & 0 & 1.9 &  & .48 & .22 & .028   & 653  \\ 
  {Base-}RSA & .41 & 0 & 2.3 &  & .48 & .22 & .027 & 659  \\ 
   \hline
\end{tabular}
\caption{Parameter fit for each model. $\lambda$ is the rationality parameter, $\Delta_{ab}$ the cost difference between \utt{A\&B} and \utt{A}, $\Delta_{a\neg b}$ the cost difference between \utt{A\&$\neg$B} and \utt{A}, and $\xi$ the extra prior used by some models (wonkiness prior in the wonky RSA model, prior on the total QUD in the supervaluationist RSA model). $\sigma_{a}$ and $\sigma_{ab}$ are the noise parameters for the comprehension data, $\epsilon$ the noise parameter for the production data. Models are sorted by AIC.}
\label{tab:free_costs_table}
\end{table}

\subsubsection{Posthoc analysis: equal costs for \utt{A\&B} and \utt{A\&$\neg$B}}

\begin{figure}[p]
\centering
\begin{subfigure}{.95\textwidth}
\centering
\includegraphics[width=\textwidth]{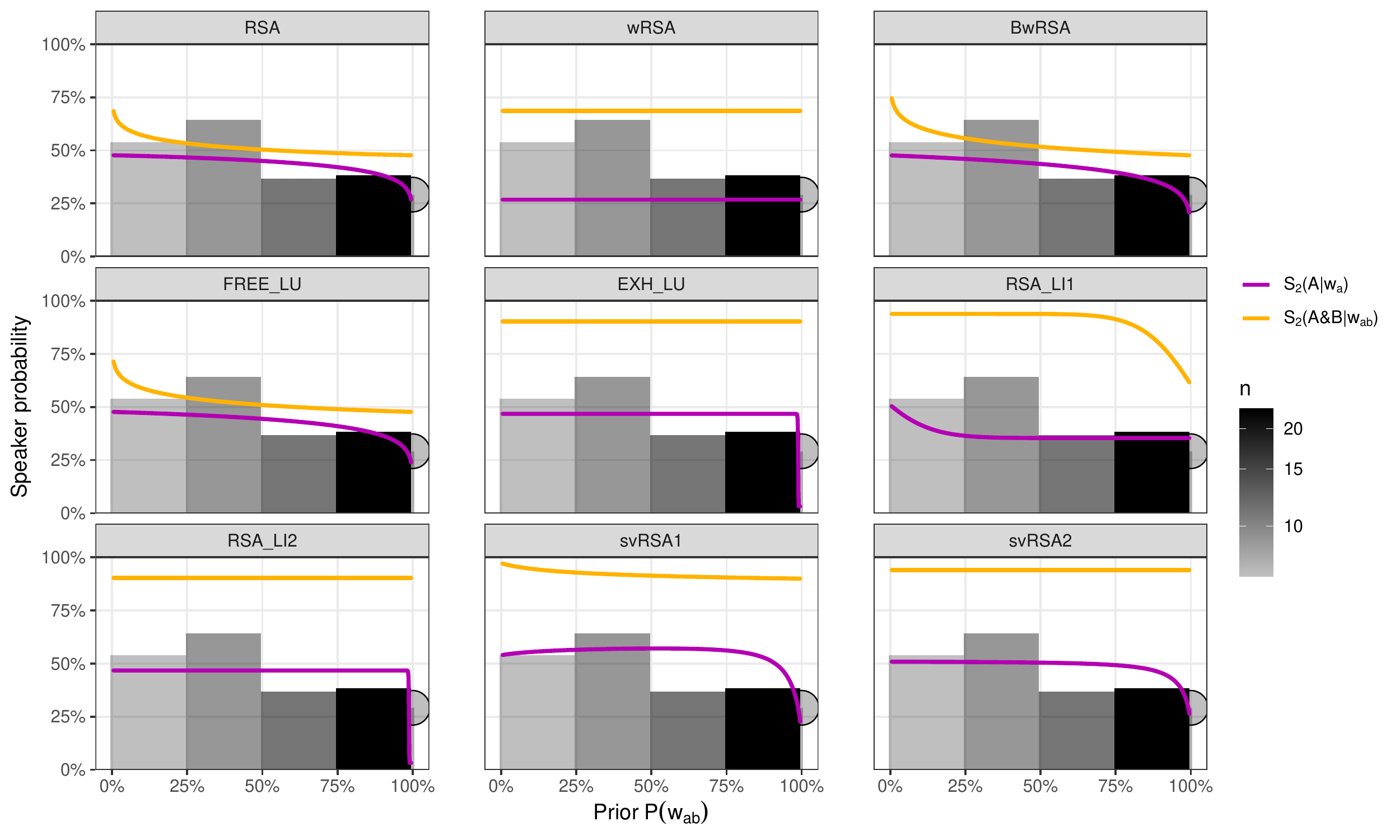}
\caption{Model fits on the production data. We only present the data from the $w_a$ world as the production for the $w_{ab}$ world consisted only of \utt{A\&B} (i.e., the orange line should be as high as possible).}
\end{subfigure}\\
\begin{subfigure}{.95\textwidth}
\centering
\includegraphics[width=\textwidth]{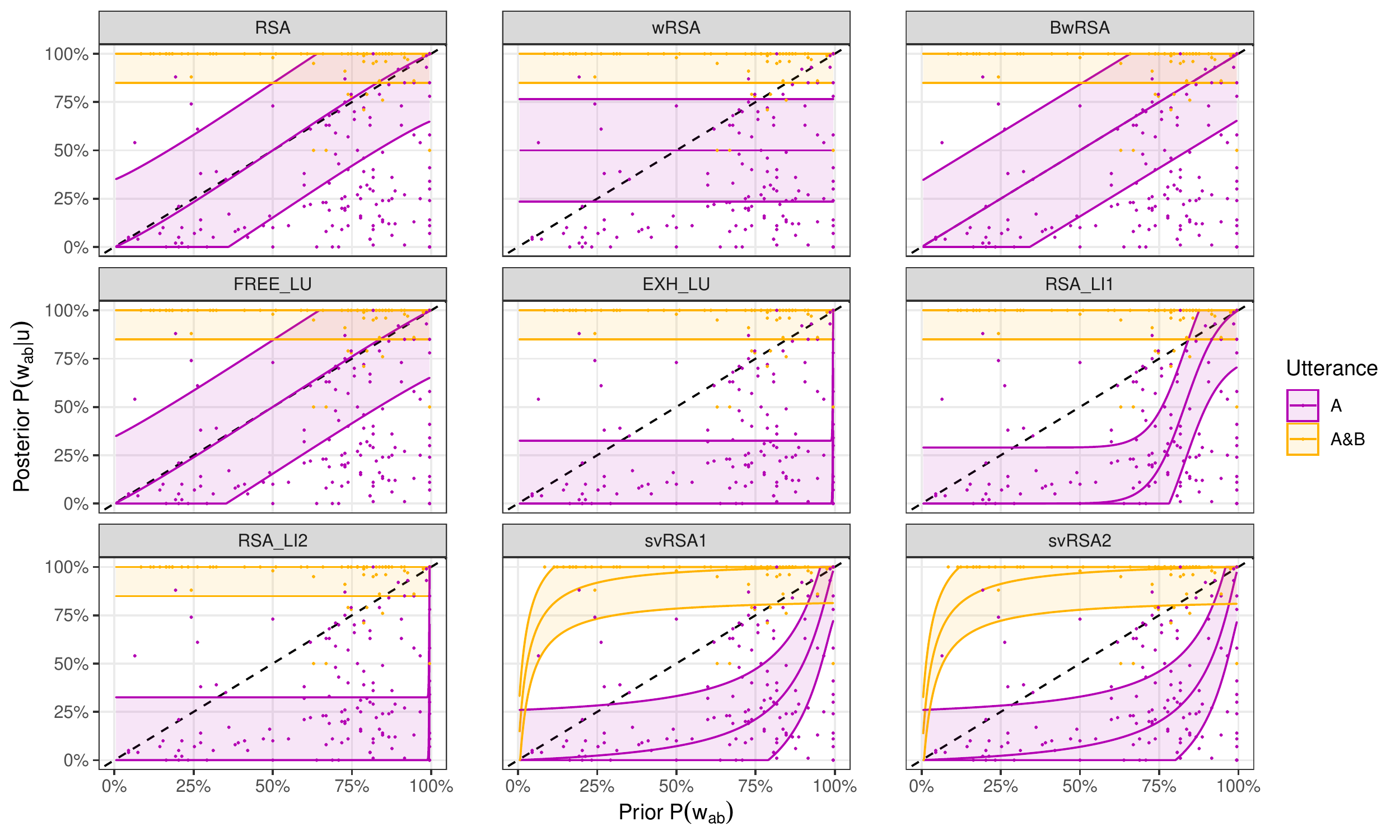}
\caption{Model fits on the comprehension data. We represent the median and quartile predictions (given the censored gaussian model for noise, the theoretical predictions correspond to the median, not the mean).}
\end{subfigure}
\caption{Model fits on production and comprehension after imposing equal costs for \utt{A\&B} and \utt{A\&$\neg$B}.}
\label{fig:equal_costs_fit}
\end{figure}

Imposing equal costs for  \utt{A\&B} and \utt{A\&$\neg$B} immediately penalizes the four models which cannot break symmetry within the semantics (wRSA, BwRSA, FREE-LU and base RSA), which fall to the bottom of Table~\ref{tab:equal_costs_table}. Undesirable behavior is evident on both the speaker and listener side, shown in Figure~\ref{fig:equal_costs_fit}: On the production side, the probability of the \utt{A\&B} message (i.e.\ the orange line) is now {clearly} below 1, meaning that these models predict the use of \utt{A} in the $w_{ab}$ world. On the comprehension side, three of them now predict the listener to just guess (i.e.\ the posterior is equal to the prior along the dashed line). By contrast, the EXH-LU, RSA-LI1, and RSA-LI2 already assigned these two messages the same null costs, so they are unaffected by this constraint. The two versions of the supervaluationist model (svRSA1 and svRSA2) remain better than the rest, even though they don't perform as well as with the extra cost on \utt{A\&B}. In particular, the prediction discussed in footnote~\ref{fn:Spector-conjunction} surfaces: when the prior on $w_{ab}$ falls sufficiently lower than the prior on the partial QUD, \utt{A\&B} can become a good message to convey the partial QUD even in $w_a$. This is visible in the production curve of svRSA1 for message \utt{A} in $w_a$---which is not monotone as \utt{A} competes with \utt{A\&B} on the left and \utt{A\&$\neg$B} on the right---, but not in that of svRSA2, as the latter assumes the total QUD for production. While some \utt{A\&B} messages do show up in the $w_a$ condition of the production survey, they do not appear to come from particularly low prior values (means: $0.70$ vs.\ $0.72$ for other messages, $t(9.6)=-.26, p=.80$), and are more likely due to errors. The problematic prediction is most visible in its effect on comprehension, as \utt{A\&B} is predicted to be compatible with $w_a$, although it never reduces the posterior below the prior (i.e., \utt{A\&B}  is never predicted to \emph{convey} $w_a$). Quantitatively, this problematic prediction does not penalize the fit of the model much, since there are very few comprehension data points for \utt{A\&B} at low prior values.

\begin{table}[ht]
\centering
\begin{tabular}{|c|c|c|c|c|c|l|c|}
  \hline
\multicolumn{1}{|c|}{Model} & \multicolumn{1}{c|}{$\lambda$} & \multicolumn{1}{c|}{$\Delta$} & \multicolumn{1}{c|}{$\xi$} & \multicolumn{1}{c|}{$\sigma_a$} & \multicolumn{1}{c|}{$\sigma_{ab}$} & \multicolumn{1}{c|}{$\epsilon$} & \multicolumn{1}{c|}{AIC}  \\ 
  \hline
svRSA1 & .85 & .41 & .91 & .38 & .28 & .003  & 514  \\ 
  svRSA2 & .68 & .18 & .91 & .39 & .28 & .022 & 529  \\ 
  RSA-LI1 &  10 & .013 &  & .43 & .22 & .023  & 555 \\ 
  RSA-LI2 & 1.0e+3 & 0 &  & .48 & .22 & .037  & 591  \\ 
  EXH-LU & 1.0e+3 & 0 &  & .48 & .22 & .037  & 591 \\ 
  wRSA & 1.4 &6.3e-4 &   1 & .39 & .22 & .026& 591  \\ 
  BwRSA & .26 & 0 &   1 & .51 & .22 & .026  & 670  \\ 
  FREE-LU & .21 & 0 &  & .51 & .22 & .026  & 671 \\ 
  RSA & .17 & 0 &  & .52 & .22 & .026 & 674 \\ 
   \hline
\end{tabular}
\caption{Parameter fit for each model when imposing equal costs for \utt{A\&B} and \utt{A\&$\neg$B}. $\lambda$ is the rationality parameter, $\Delta$ the cost difference between \utt{A\&B}/\utt{A\&$\neg$B} and \utt{A}, and $\xi$ the extra prior used by some models (wonkiness prior in the wonky RSA, prior on the total QUD in the supervaluationist model). $\sigma_{a}$ and $\sigma_{ab}$ are the noise parameters for the comprehension data, $\epsilon$ the noise parameter for the production data. Models are sorted by AIC.}
\label{tab:equal_costs_table}
\end{table}

\section{Discussion}

{By measuring priors in the same experiment as target behavior, our design} improved on previous studies and allowed us to detect the effect of priors on production, which had been predicted but not established. We also found an effect of priors on comprehension, with an effect size almost 8x larger than \cite{degen2015wonky} when it comes to the correlation between priors and posteriors ($\beta=.49$ vs $\beta=.06$), yet we still do not find any trace of anti-exhaustivity.\footnote{\cite{javangula2019} did find an even stronger correlation than we did: $\beta=.79$, albeit in a more restricted range of priors, from 0.30 to 0.80 prior conditional probability for the $w_{ab}$ world, where we would not expect anti-exhaustivity anyway.}

At the highest level, our results pose a challenge to simple information-theoretic models of pragmatics, like the base RSA model, insofar as they are not compatible with {either} production and comprehension being {at the same time} rational and well-informed. If our production data is taken as the basis for RSA-style comprehension, the participants are not rational, in a Bayesian sense. In this respect, our conclusions align with \cite{sikos2021reevaluating}, who also found that listeners don't align with speakers in a recent replication of \pocite{frank2012predicting} experiment.\footnote{RSA models do not generally predict that the listener and the speaker will be rational with respect to each other: if we use $L_1$ and $S_1$ as models for the listener and speaker, then the speaker is assumed to be (approximately) rational with respect to  $L_0$, not $L_1$. If instead we take $L_0$ to be the model for the listener and $S_1$ for the speaker, then we do not model the listener as being rational with respect to the speaker. Yet RSA models do assume that at least one of the two participants is rational with respect to the other: with $L_1$ and $S_1$, the listener is modeled as rational with respect to the speaker (but not vice-versa). With $L_0$ and $S_1$, or $L_1$ and $S_2$, the speaker is modeled as (approximately) rational with respect to the listener, but not vice-versa.}

 Likewise, if our comprehension data is taken as the basis for utterance production, then anti-exhaustivity effects can only be avoided with implausible costs. Our results, however, do not exclude models in which speakers and listeners are rational but do not share the same information (e.g., knowledge of extra parameters which can be introduced in the RSA framework), or models which depart from the rationality assumption for one of the agents (e.g., by allowing the listener to revise her priors, as in some implementations of ``wonky world'' RSA). Lastly, we found that models which can break symmetry in the semantics by implementing grammatical exhaustification have a clear advantage over models which must rely on informativity and costs alone (see \citealp{champollion2019free} and \citealp{franke2020theory}  for similar conclusions from different phenomena). 

The EXH-LU and LI2 models converged to parameters that turn them into a categorical model: null costs and extremely high rationality mean that the speaker always selects the most informative message. In turn the listener can retrieve the world with near absolute certainty, so priors play no role on the interpretation side.

The supervaluationist model (svRSA) emerged as the best candidate to model our data set. It appears to be particularly good at capturing speaker/listener asymmetries thanks to its QUD parameter, and its approach to ambiguity resolution allows it to take full advantage of exhaustification. Unlike the LI model, in which ambiguity offers the speakers more opportunities to use a given message, the supervaluationist model restricts the use of ambiguous messages to situations where they are true under all possible interpretations. This constraint blocks anti-exhaustivity in both production and comprehension, independently of costs parameters, and can only be relaxed when a partial QUD makes exhaustivity irrelevant. The model does however make a problematic prediction regarding the use of \utt{A\&B} to convey the pair ($w_a$, partial QUD) when priors become too extreme and the cost of \utt{A\&B} too low.

Like previous studies, our experiment used written stimuli in English. Since English usually marks focus prosodically, this leaves the  focus structure of our message \utt{A} underspecified. Many of the models we tested assumed ambiguity, but focus and covert exhaustification are usually assumed to correlate, so prosody could contribute to disambiguation, and this could take many forms. First, we could assume that prosody fully disambiguates the intended meaning, so that unfocused \utt{A} is obligatorily literal and focused \utt{A} obligatorily exhaustive. Second, we could assume that the exhaustification operator requires focus, so that unfocused \utt{A} is unambiguously literal, but focused \utt{A} would remain ambiguous. Conversely, we could assume that focus requires exhaustification, so that  focused \utt{A} is obligatorily exhaustive but unfocused \utt{A} is now ambiguous. The problem becomes even more complex if we consider these connections to be probabilistic rather than deterministic. Alternatively, the connection between focus and exhaustification could be indirect, for instance if focus conveys information about the QUD.

One reason why previous investigations may have underestimated {the issue of} anti-exhaustivity in RSA models is that model predictions are often derived without taking into account explicit exhaustive messages such as \utt{A and not B}, relying instead on a predefined set of alternatives defined along the lines of standard neo-Gricean approach (as discussed in footnote \ref{alternativesfn}). This is the case in both \cite{degen2015wonky} and \cite{potts2016embedded}, which tested wRSA and RSA-LU respectively. Doing so amounts to artificially breaking the symmetry by assigning an infinite cost to this message, which prevents anti-exhaustivity even in the base RSA model. Yet, our production data show that such messages can easily be expressed with \utt{only} or \utt{just} and are produced as often as the weak message \utt{A}, so they shouldn't be ignored in modeling. Furthermore, we did detect the expected influence of priors on message choice when the the intended meaning is $w_a$. As predicted by virtually all RSA models, participants are more likely to use an explicit message whose meaning is $A\wedge\neg B$ when conditional probability of $A\&B$ given $A$ is very high. The trade-off between cost and informativity in message choice, which is a key feature of RSA models with cost, is a virtue (see also {\citealp{chemla2015probability} and } \citealp{enguehard2021explaining}, which include clear introspective judgments suggesting that the choice between \emph{some} and \emph{not all} is sensitive to priors in the expected way), but it appears that RSA models, to be able to predict such effects without predicting at the same time anti-exhaustivity effects, need to include exhaustification in their semantic component.

\citet{fox2021notes} have previously discussed the fact that probabilistic models of scalar implicatures and exhaustivity effects (such as models couched within the RSA framework) let us expect a strong influence of prior probabilities on such phenomena -- cases where different priors should lead to drastically different interpretations for the very same sentence. They argue that, if in fact such an influence is not found in these domains, such a finding should shed doubt on pragmatic theories of exhaustivity that rely on such models. They point out that even if some relatively complex probabilistic models manage to eliminate the undesirable effect of priors, their success would then not be due to what is \emph{distinctive} about probabilistic approaches.  Our results are consistent with this view, since they provide new evidence that exhaustive interpretations should be derived within the compositional \emph{semantic} system rather than being viewed as a purely \emph{pragmatic} phenomenon (in line with some other recent studies in the RSA literature, e.g., \citealt{franke2020theory}). The relevant semantic mechanism is indeed completely encapsulated from  probabilistic reasoning. Now, precisely because this mechanism is not a pragmatic mechanism but rather belongs to compositional semantics, this conclusion is by itself fully compatible with the view that pragmatic reasoning proper is probabilistic in nature and, specifically, that the notion of informativity relevant to pragmatics is the information-theoretic notion of \emph{surprisal} used in the RSA framework. Our data, specifically on the production side, suggest that this is in fact the case. In a semantic theory that includes an exhaustivity mechanism, sentences are typically ambiguous between an exhaustive and a non-exhaustive reading, and so an account of the use and interpretation of potentially ambiguous sentences needs to include an explicit formalization of how speakers and listeners deal with this ambiguity. The RSA framework provides the means to articulate a semantic view of exhaustivity with such a formally explicit pragmatic model, in which the role of prior probabilistic expectations can be captured. 

\clearpage
\bibliographystyle{apacite}

\bibliography{CremersWilcoxSpector}

\pagebreak

\appendix

\section{Formal conditions leading to Exhaustivity and Anti-Exhaustivity}

\subsection{Baseline RSA model}
\label{sec:RSA-antiexh-proof}

Consider the baseline RSA model with two worlds, $w_a$ and $w_{ab}$ and three messages,  \utt{A}, $A\wedge B$ and $A \wedge \neg B$. Let us set the cost of \utt{A} at $0$, and notate $c_{A\wedge B}$  and $c_{A \wedge \neg B}$ the costs of the two other messages.As usual, $\lambda$ is a positive real number, and $P$ is the prior probabilty distribution over the two worlds.\\
\\
We will first prove the following resuts:

\ex.
\a. $S_1(A|w_{ab}) > S_1(A|w_a))$ if and only if $\log(P(w_{ab})) - \log(P(w_a)) > c_{A \wedge \neg B} - c_{A \wedge B}$
\b. $L_1(w_{ab}|A) > P(w_{ab}|\{w_a, w_{ab}\})$ if and only if $\log(P(w_{ab})) - \log(P(w_a)) > c_{A \wedge \neg B} - c_{A \wedge B}$

Let us note that in this simple model, since there are only two worlds, $P(w_{ab}|\{w_a, w_{ab}\}) = P(w_{ab})$, and the message \utt{A} is in fact the tautology. For this reason, for either world $w$ in $\{w_a, w_{ab}\}$,  $L_0(w|A) = P(w)$. 

Given the equations of the RSA model, and the observations just made, we have:

\ex. 
\a. $U_1(A|w_a) = \log(P(w_A))$
\b. $U_1(A \wedge \neg B|w_A) = \log(\underbrace{L_0(w_a|A \wedge \neg B)}_1) - c_{A \wedge \neg B} = - c_{A \wedge \neg B}$
\c. $U_1(A|w_{ab}) = \log(P(w_{ab}))$
\d. $U_1(A \wedge B|w_{ab}) = \log(\underbrace{L_0(w_{ab}|A \wedge B)}_1) - c_{A \wedge  B} = - c_{A \wedge  B}$

Given this, applying the definitions, we have:

\ex. $S_1(A|w_a) = \dfrac{e^{\lambda\log(P(w_a))}}{e^{\lambda\log(P(w_a))} + e^{- \lambda c_{A \wedge \neg B}}} = \dfrac{1}{1 + e^{-\lambda(c_{A\wedge \neg B} +\log(P(w_a))}}$

Likewise:

\ex. $S_1(A|w_{ab}) = \dfrac{e^{\lambda\log(P(w_{ab}))}}{e^{\lambda\log(P(w_{ab}))} + e^{- \lambda c_{A \wedge B}}} = \dfrac{1}{1 + e^{-\lambda(c_{A\wedge B} +\log(P(w_{ab}))}}$

Since the {logistic} function $\frac{1}{1+e^{-\lambda x}}$ is increasing, $$S_1(A|w_{ab})~>~S_1(A|w_a)\textrm{~iff~} c_{A\wedge~B}+\log(P(w_{ab}))~>~c_{A\wedge\neg~B}+\log(P(w_{a}))$$

\noindent This is in turn equivalent to:
$$\log(P(w_{ab})) - \log(P(w_a)) > c_{A \wedge \neg B} - c_{A \wedge B}$$

\noindent Let us now prove that: 

\ex. $L_1(w_{ab}|A) > P(w_{ab})$ iff $S_1(A|w_{ab}) > S_1(A|w_a)$

We have:

$$L_1(w_{ab}|A) = \dfrac{P(w_{ab})\times S_1(A|w_{ab})}{P(w_{ab})\times S_1(A|w_{ab}) + P(w_a)\times S_1(A|w_a)}$$\\
 
\noindent With $p = P(w_{ab})\in(0,1)$, we have:

\begin{align*}
L_1(w_{ab}|A)  > P(w_{ab})
& \equiv &  \dfrac{p \times S_1(A|w_{ab})}{p \times S_1(A|w_{ab}) + (1-p)\times S_1(A|w_a)} > p\\
& \equiv &\dfrac{S_1(A|w_{ab})}{p\times S_1(A|w_{ab}) + (1-p)\times S_1(A|w_a)} > 1\\
& \equiv & S_1(A|w_{ab}) > p\times S_1(A|w_{ab}) + (1-p)\times S_1(A|w_a)\\
& \equiv & (1-p)S_1(A|w_{ab}) > (1-p) S_1(A|w_a)\\
& \equiv & S_1(A|w_{ab}) > S_1(A|w_a)
\end{align*}

\subsection{Wonky RSA models}
\label{sec:wRSA-antiexh-proof}

In this model, the $S_1$ speaker is conditionalized on the wonkiness parameter, and therefore not interpretable as an actual speaker. We therefore consider anti-exhaustivity at the $L_1$ listener level only. The full expression of $L_1$ in the standard wRSA and our Bayesian implementation BwRSA are given in \Next[a] and \Next[b] respectively, where $p=P(w_{ab})$ is the non-wonky prior on the world which makes \utt{A and B} true, $\omega$ is the wonkiness prior, $\lambda$ the rationality parameter, and $c_{A \wedge  B}$ and $c_{A \wedge\neg  B}$ the costs of the \utt{A and B} and \utt{A and not B} messages, respectively. $f_\lambda$ represents the logistic function with rate $\lambda$.

\ex.
	\a. {\footnotesize $\displaystyle L_1(w_{ab}|A) = \frac{p(1-\omega)f_\lambda(\log p + c_{A \wedge  B})+\frac12 \omega f_\lambda(c_{A \wedge  B}-\log 2)}{p(1-\omega)f_\lambda(\log p + c_{A \wedge  B})+\frac12 \omega f_\lambda(c_{A \wedge  B}-\log 2) + (1-p)(1-\omega)f_\lambda(\log (1-p) + c_{A \wedge\neg  B}) + \frac12 \omega f_\lambda(c_{A \wedge\neg  B}-\log 2)}$}
	\b. {\footnotesize $\displaystyle L_1(w_{ab}|A) = p\frac{(1-\omega)f_\lambda(\log p + c_{A \wedge  B})+\omega f_\lambda(c_{A \wedge  B}-\log 2)}{p[(1-\omega)f_\lambda(\log p + c_{A \wedge  B})+\omega f_\lambda(c_{A \wedge  B}-\log 2)] + (1-p)[(1-\omega)f_\lambda(\log (1-p) + c_{A \wedge\neg  B}) +  \omega f_\lambda(c_{A \wedge\neg  B}-\log 2)]}$}

In both cases, we define listener-side anti-exhaustivity as any situation where the posterior exceeds the measured prior, i.e.\ $L_1(w_{ab}|A) > P(w_{ab}|b_1)$.

For the wRSA, an analytical solution would be difficult to express, but we can already make two observation: if $\omega>0$, the model always predicts anti-exhaustivity when $p\to0$, and always predicts exhaustivity when $p\to1$. This is just a consequence of it being non-Bayesian, as the posterior value never converges to exactly 0 and 1, respectively. The anti-exhaustivity when $p\to0$ is unrelated to anti-exhaustivity in the baseline RSA, which happens when $p$ is high, and is arguably unproblematic. The main question is whether anti-exhaustivity can arise in the wRSA when $p$ is high but not too close to 1. With numerical simulations, we can answer positively. For instance, if we set the parameters to $\omega=0.1, c_{A \wedge  B}=1,c_{A \wedge \neg B}=1.2,\lambda=3$, we find anti-exhaustivity for $p<.04$ but also for $p\in[0.57,0.95]$. The problematic anti-exhaustivity inherited from the baseline RSA fades away with sufficient increase in either $\omega$ or $c_{A \wedge \neg B}$.

For the Bayesian version BwRSA, we can show that anti-exhaustivity happens if and only if:
\[(1-\omega)\left[f_\lambda(\log p +c_{A \wedge  B}) - f_\lambda(\log (1-p) +c_{A \wedge\neg  B})\right]+
\omega\left[f_\lambda(c_{A \wedge  B}-\log 2) - f_\lambda(c_{A \wedge\neg  B}-\log 2)\right] > 0\]
Since this is a monotone increasing function of $p$, we can simply look at the situation when $p\to1$, and it follows that a sufficient and necessary condition on the presence of anti-exhaustivity in this model is:
\[\omega < \frac{f_\lambda(c_{A \wedge  B})}{f_\lambda(c_{A \wedge  B}) - f_\lambda(c_{A \wedge  B}-\log 2) + f_\lambda(c_{A \wedge\neg  B} - \log 2)}\]
In practice, this means that anti-exhaustivity is very difficult to avoid in this model. Numerically, $\omega$ would have to be over 0.9 with the parameter values used in the wRSA case.

\subsection{Supervaluationist model}
\label{sec:Spector-antiexh-proof}

At the level of $S_1$, the model predicts no anti-exhaustivity in the sense that for any QUD $Q$, no matter the priors, $S_1(A|w_{ab},Q) \leq S_1(A|w_a,Q)$. Indeed, if $Q$ distinguishes between $w_a$ and $w_{ab}$, the fact that \utt{A} can be interpreted as $exh(A)$ entails $S_1(A|w_{ab},Q)=0$, while if $Q$ does not distinguish between these worlds, $S_1(A|w_{ab},Q) = S_1(A|w_a,Q)$.

Turning to the pragmatic listener $L_1$, we can prove that the posterior on $w_{ab}$ after hearing \utt{A} never exceeds the prior (with equality only if the prior is exactly 0 or 1).

\[\begin{array}{rcl}
L_1(w_{ab}|A) &=& L_1(w_{ab}|A,Q_A) + L_1(w_{ab}|A,Q_\text{total})\\
L_1(w_{ab}|A) &=& \displaystyle \frac{P(w_{ab})P(Q_A)S_1(A|w_{ab},Q_A)+P(w_{ab})P(Q_\text{total})\overbrace{S_1(A|w_{ab},Q_\text{total})}^{=0}}{\sum_{w,Q} P(w)P(Q)S_1(A|w,Q)}\\
&&\rule[-3ex]{0pt}{6ex}\hspace{-2cm}\text{Noting that }S_1(A|w_{a},Q_A)=S_1(A|w_{ab},Q_A)=S_1(A|Q_A)\text{ we can factorize the denominator and simplify:}\\
L_1(w_{ab}|A) &=&  \displaystyle \frac{P(w_{ab})P(Q_A)S_1(A|Q_A)}{(P(w_{a}) + P(w_{ab}))P(Q_A)S_1(A|Q_A) + P(w_{a})P(Q_\text{total})S_1(A|w_{a},Q_\text{total})}\\
L_1(w_{ab}|A) &=&  \displaystyle \frac{P(w_{ab})}{\displaystyle1 + P(w_{a})\underbrace{\frac{P(Q_\text{total})S_1(A|w_{a},Q_\text{total})}{P(Q_A)S_1(A|Q_A)}}_{>0}} \leq P(w_{ab})\\
\end{array}
\]

\section{Implementation of the models}
\label{app:implementations}

Here we list the formulas used to implement the different models. See the file ``RSAExh-ModelFunctions.R'' in supplementary materials for details on how we translated these formulas to \texttt{R}. In all models, $\lambda$ represents the rationality parameter, $\Delta_{ab}$ and $\Delta_{a\neg b}$ the costs of \utt{A\&B} and \utt{A\&$\neg$B} relative to \utt{A}, and $p$ the conditional prior on $w_{ab}$ given $\{w_a,w_{ab}\}$. Since we consider only two worlds, we only give marginal listener functions for $w_{ab}$. The listener function for $w_a$ is simply the complementary. 

\subsection{Base RSA}

$L_1$ listener:
\[\begin{array}{rcl}
	L_1(w_{ab}|A) &=& \frac{1}{1+ \frac{(1-p)\left(1+\mathrm{e}^{-\lambda(\log(p)+\Delta_{ab})}\right)}{p\left(1+ \mathrm{e}^{-\lambda(\log(1-p)+\Delta_{a\neg b})}\right)} }\\
	L_1(w_{ab}|A\&B) &=& 1\\
	L_1(w_{ab}|A\&\neg B) &=& 0
\end{array}\]

\noindent$\log S_2$ speaker:
\[\begin{array}{rcl}
	\log S_2(A|w_a) &=& -\log\left(1+\mathrm{e}^{-\lambda(\log(1-L_1(w_{ab}|A))+\Delta_{a\neg b})}\right)\\
	\log S_2(A\&\neg B|w_a) &=& -\log\left(1+\mathrm{e}^{\lambda(\log(1-L_1(w_{ab}|A))+\Delta_{a\neg b})}\right)\\	
	\log S_2(A|w_{ab}) &=& -\log\left(1+\mathrm{e}^{-\lambda(\log L_1(w_{ab}|A)+\Delta_{ab})}\right)\\	
	\log S_2(A\&B|w_{ab}) &=& -\log\left(1+\mathrm{e}^{\lambda(\log L_1(w_{ab}|A)+\Delta_{ab})}\right)
\end{array}\]

\subsection{wRSA and BwRSA}

The $S_1$ is parametrized by the wonkiness parameter $b$ which can take values $b_1$ (wonky) or $b_2$ (usual), and a parameter $\omega=P(b=b_1)$ which is the prior on wonkiness. The ``ususal'' $S_1$ speaker is the same as the base RSA speaker. The ``wonky'' $S_1$ speaker is given by:
\[\begin{array}{rcl}
	 S_1(A|w_a,b_1) &=& \frac{1}{1+\mathrm{e}^{\lambda(\log 2-\Delta_{a\neg b})}}\\
	 S_1(A\&\neg B|w_a,b_1) &=& \frac{1}{1+\mathrm{e}^{-\lambda(\log 2-\Delta_{a\neg b})}}\\
	 S_1(A|w_{ab},b_1) &=& \frac{1}{1+\mathrm{e}^{\lambda(\log 2-\Delta_{ab})}}\\	
	 S_1(A\&B|w_{ab},b_1) &=& \frac{1}{1+\mathrm{e}^{-\lambda(\log 2-\Delta_{ab})}}
\end{array}\]

For wRSA, the $L_1$ speaker updates her prior to reflect her assumptions about the speaker. Here we consider the marginal posterior on the world.
\[\begin{array}{rcl}
	L_1(w_{ab}|A) &=& \frac{\frac12 \omega S_1(A|w_a,b_1) + p (1-\omega)S_1(A|w_a,b_2)}{
		\frac12 \omega \left(S_1(A|w_a,b_1) +  S_1(A|w_{ab},b_1)\right) +
		(1-\omega) (p S_1(A|w_a,b_2) + (1-p)S_1(A|w_{ab},b_2))}
\end{array}\]

In the BwRSA, the $L_1$ speaker makes inferences about the speaker's priors, but sticks to her own priors when inferring $w$. Again, we consider the marginal posterior on the world:
\[\begin{array}{rcl}
	L_1(w_{ab}|A) &=& p\frac{\omega S_1(A|w_a,b_1) + (1-\omega)S_1(A|w_a,b_2)}{
		\omega \left(p S_1(A|w_a,b_1) +   (1-p) S_1(A|w_{ab},b_1)\right) +
		(1-\omega) (p S_1(A|w_a,b_2) + (1-p)S_1(A|w_{ab},b_2))}
\end{array}\]

In both cases, the strong messages are interpreted literally: $L_1(w_{ab}|A\&B) = 1$ and $L_1(w_{ab}|A\&\neg B) = 0$. At the level of $S_2$, the parameter $b$ does not appear, and the speaker only tries to convey the world, so the formulas are identical to base RSA.

\subsection{svRSA1 and svRSA2}

Here we consider two possible QUDs: $Q_A$ is  the partial QUD and corresponds to the partition $\{\{w_a,w_{ab}\}\}$ (which is trivial when considering only two worlds) and $Q_{total}$ is the total QUD and corresponds to the partition $\{\{w_a\},\{w_{ab}\}\}$. We write $q=P(Q=Q_{total})$ the prior on QUDs. We write $\chi=P(i_{exh})$ the prior on exhaustification, which will be fixed to $\frac12$ in the model fit.

For $Q_A$, the world does not affect the choice of a message since $Q_A(w_a)=Q_A(w_{ab})$, and only costs distinguish the three messages.
\[\begin{array}{rcl}
	 S_1(A|w,Q_A) &=& \frac{1}{1+\mathrm{e}^{-\lambda\Delta_{a\neg b}}+\mathrm{e}^{-\lambda\Delta_{ab}}}\\
	 S_1(A\&\neg B|w,Q_A) &=& \frac{1}{1+\mathrm{e}^{\lambda\Delta_{a\neg b}}+\mathrm{e}^{\lambda(\Delta_{a\neg b}-\Delta_{ab})}}\\
	 S_1(A\&B|w,Q_A) &=& \frac{1}{1+\mathrm{e}^{\lambda\Delta_{ab}}+\mathrm{e}^{\lambda(\Delta_{ab}-\Delta_{a\neg b})}}\\
\end{array}\]	

For $Q_{total}$, the world and therefore the meaning of the messages matter:
\[\begin{array}{rcl} 
	S_1(A|w_a,Q_{total}) &=& \frac{1}{1+ \mathrm{e}^{-\lambda((1-\chi)\log(1-p)+\Delta_{a\neg b})}}\\
	S_1(A\&\neg B|w_a,Q_{total}) &=& \frac{1}{1+ \mathrm{e}^{\lambda((1-\chi)\log(1-p)+\Delta_{a\neg b})}}\\
	S_1(A|w_{ab},Q_{total}) &=& 0\\
	S_1(A\&B|w_{ab},Q_{total}) &=& 1
\end{array}\]
Note that under the total QUD, \utt{A} can only be used to convey $w_a$, so only \utt{A\&B} can convey~$w_{ab}$.

$L_1$ infers both the QUD and the world. We give the joint posterior distribution here (which is needed to derive $S_2$), but the fit of the comprehension data was based on the marginal world posterior.
\[\begin{array}{rcl}
	L_1(w_{a},Q_A|A) &=& \frac{(1-p)(1-q)S_1(A|w,Q_A)}{(1-q)S_1(A|w,Q_A) + q(1-p)S_1(A|w_a,Q_{total})}\\
	L_1(w_{a},Q_{total}|A) &=& \frac{(1-p)qS_1(A|w_a,Q_{total})}{(1-q)S_1(A|w,Q_A) + q(1-p)S_1(A|w_a,Q_{total})}\\
	L_1(w_{ab},Q_A|A) &=& \frac{p(1-q)S_1(A|w,Q_A)}{(1-q)S_1(A|w,Q_A) + q(1-p)S_1(A|w_a,Q_{total})}\\
	L_1(w_{a},Q_A|A\&B) &=& \frac{(1-p)(1-q)S_1(A\&B|w,Q_A)}{(1-q)S_1(A\&B|w,Q_A) + qp}\\
	L_1(w_{ab},Q_A|A\&B) &=& \frac{p(1-q)S_1(A\&B|w,Q_A)}{(1-q)S_1(A\&B|w,Q_A) + qp}\\
	L_1(w_{ab},Q_{total}|A\&B) &=& \frac{qp}{(1-q)S_1(A\&B|w,Q_A) + qp}\\
	L_1(w_{a},Q_A|A\&\neg B) &=&  \frac{(1-p)(1-q)S_1(A\&\neg B|w,Q_A)}{(1-q)S_1(A\&\neg B|w,Q_A) + q(1-p)S_1(A\&\neg B|w_a,Q_{total})}\\
	L_1(w_{ab},Q_A|A\&\neg B) &=&  \frac{p(1-q)S_1(A\&\neg B|w,Q_A)}{(1-q)S_1(A\&\neg B|w,Q_A) + q(1-p)S_1(A\&\neg B|w_a,Q_{total})}\\
	L_1(w_{a},Q_{total}|A\&\neg B) &=&  \frac{(1-p)qS_1(A\&\neg B|w_a,Q_{total})}{(1-q)S_1(A\&\neg B|w,Q_A) + q(1-p)S_1(A\&\neg B|w_a,Q_{total})}\\
\end{array}\]
All other combinations involve $Q_{total}$ and world that makes the message literally false, and therefore receive a posterior probability of~0. 

We can now give the $\log S_2$ functions. The functions for $Q_A$ only depend on the $L_1$ marginal $Q_A$ posterior, since a speaker who wants to convey $Q_A$ does not care about $w$.

\[\begin{array}{rcl}
	\log S_2(A|w,Q_A) &=& -\log\left(1+\mathrm{e}^{-\lambda\left(\log \frac{L_1(Q_A|A)}{L_1(Q_A|A\&\neg B)}+\Delta_{a\neg b}\right)}+\mathrm{e}^{-\lambda\left(\log \frac{L_1(Q_A|A)}{L_1(Q_A|A\& B)}+\Delta_{a b}\right)}\right)\\
	\log S_2(A\&\neg B|w,Q_A) &=&  -\log\left(1+\mathrm{e}^{\lambda\left(\log \frac{L_1(Q_A|A)}{L_1(Q_A|A\&\neg B)}+\Delta_{a\neg b}\right)}+\mathrm{e}^{-\lambda\left(\log \frac{L_1(Q_A|A\&\neg B)}{L_1(Q_A|A\& B)}+\Delta_{a b}-\Delta_{a\neg b}\right)}\right)\\
	\log S_2(A\&B|w,Q_A) &=& -\log\left(1+\mathrm{e}^{-\lambda\left(\log \frac{L_1(Q_A|A\&B)}{L_1(Q_A|A\&\neg B)}+\Delta_{a\neg b}-\Delta_{a b}\right)}+\mathrm{e}^{\lambda\left(\log \frac{L_1(Q_A|A)}{L_1(Q_A|A\& B)}+\Delta_{a b}\right)}\right)\\
	\log S_2(A|w_a,Q_{total}) &=& -\log\left(1+\mathrm{e}^{-\lambda\left(\log \frac{L_1(w_a,Q_{total}|A)}{L_1(w_a,Q_{total}|A\&\neg B)}+\Delta_{a\neg b}\right)}\right)\\
	\log S_2(A\&\neg B|w_a,Q_{total}) &=& -\log\left(1+\mathrm{e}^{\lambda\left(\log \frac{L_1(w_a,Q_{total}|A)}{L_1(w_a,Q_{total}|A\&\neg B)}+\Delta_{a\neg b}\right)}\right)\\
	\log S_2(A\&B|w_{ab},Q_{total}) &=& 0
\end{array}\]

In practice, we tested two implementations of the supervaluationist model for production. In svRSA1, the production was modelled as $(1-q)S_2(u|w,Q_A) + qS_2(u|w,Q_{total})$, representing uncertainty on the QUD the speaker should convey. In svRSA2 it was simply modelled as $S_2(u|w,Q_{total})$, which meant that the speaker had no uncertainty on which QUD to convey, but the listener still had to infer the QUD and the speaker was aware of the listener's uncertainty.

\subsection{FREE-LU and EXH-LU}

We first introduced a full unrestricted LU model with all possible strengthening and free priors over these, from which we will define our two implementations. We consider three possible interpretations $i_{lit},i_{exh},i_\textit{anti-exh}$ such that: 
\[\begin{array}{lccl}
	\sem{A}^{i_{lit}} &=& \{w_a,w_{ab}\} & \quad\text{literal} \\
	\sem{A}^{i_{exh}} &=& \{w_a\} & \quad\text{exhaustive} \\
	\sem{A}^{i_\textit{anti-exh}} &=& \{w_{ab}\} & \quad\text{anti-exhaustive}
\end{array}\]
For messages \utt{A\&B} and \utt{A\&$\neg$B}, all $i$'s return the literal interpretations, as they cannot be further strengthened.

\[\begin{array}{rcl}
	 S_1(A|w_a,i_{lit}) &=& \frac{1}{1+\mathrm{e}^{-\lambda(\log(1-p) + \Delta_{a\neg b})}}\\
	 S_1(A|w_{ab},i_{lit}) &=& \frac{1}{1+\mathrm{e}^{-\lambda(\log p + \Delta_{ab})}}\\
	 S_1(A|w_a,i_{exh}) &=& \frac{1}{1+\mathrm{e}^{-\lambda\Delta_{a\neg b}}}\\
	 S_1(A|w_{ab},i_\textit{anti-exh}) &=& \frac{1}{1+\mathrm{e}^{-\lambda\Delta_{ab}}}
\end{array}\]	

Moving to $L_1$, we need to consider the priors on these three interpretations $\rho_{lit},\rho_{exh},\rho_\textit{anti-exh}$. The posteriors for $w_{ab}$ after \utt{A\&B} and \utt{A\&$\neg$B} remain 1 and 0, respectively.
\[
L_1(w_{ab}|A) = p\frac{\rho_{lit}S_1(A|w_{ab},i_{lit})+\rho_\textit{anti-exh} S_1(A|w_{ab},i_\textit{anti-exh})}{(1-p)\rho_{lit}S_1(A|w_{a},i_{lit})+p\rho_{lit}S_1(A|w_{ab},i_{lit})+(1-p)\rho_{exh}S_1(A|w_{a},i_{exh})+p\rho_\textit{anti-exh} S_1(A|w_{ab},i_\textit{anti-exh})}
\]

The $S_2$ speaker is defined as in baseline RSA. We derive the FREE-LU model by fixing $\rho_{lit}=\rho_{exh}=\rho_\textit{anti-exh}=\frac13$ and the EXH-LU model by fixing $\rho_{lit}=\rho_{exh}=\frac12$ and $\rho_\textit{anti-exh}=0$.

\subsection{RSA-LI}

For the RSA-LI model, we keep interpretations $i_{lit}$ and $i_{exh}$ from the LU model, but the speaker now tries to convey which interpretation they meant. There is no prior on $i$ anymore (in other words, the listener has a uniform prior on $i$). We can look directly at the marginal $S_1$ speaker.
\[\begin{array}{rcl}
	 S_1(A|w_a) &=& \frac{1+\mathrm{e}^{\lambda\log(1- p)}}{1+\mathrm{e}^{\lambda\log(1- p)} + 2\mathrm{e}^{-\lambda\Delta_{a\neg b}}}\\
	 S_1(A|w_{ab}) &=& \frac{1}{1+2\mathrm{e}^{-\lambda(\log p + \Delta_{ab})}}\\
	 S_1(A\&\neg B|w_a) &=& \frac{2}{2+\mathrm{e}^{\lambda\left(\log(1- p)+\Delta_{a\neg b}\right)} + \mathrm{e}^{\lambda\Delta_{a\neg b}}}\\
	 S_1(A\&B|w_{ab}) &=& \frac{2}{2+\mathrm{e}^{\lambda(\log p + \Delta_{ab})}}
	 
\end{array}\]

The marginal $L_1$ listener is then described by:
\[\begin{array}{rcl}
	L_1(w_{ab}|A) &=& \frac{1}{1 + \frac{(1-p)\left(1+2\mathrm{e}^{-\lambda(\log p + \Delta_{ab})}\right)\left(1+\mathrm{e}^{\lambda\log(1- p)}\right)}{p\left(1+\mathrm{e}^{\lambda\log(1- p)} + 2\mathrm{e}^{-\lambda\Delta_{a\neg b}}\right)}}
\end{array}\]

The $S_2$ speaker is defined as the usual RSA speaker (she cannot select a specific meaning, as $L_1$---unlike $L_0$---is assumed to be blind to the intended meaning). Because of this crucial difference between $S_1$ and $S_2$, we tested both speakers as models of production ($S_1$ in RSA-LI1 and $S_2$ in RSA-LI2).

\end{document}